\documentclass[runningheads]{llncs}

\usepackage{eccv}

\usepackage{eccvabbrv}

\usepackage{graphicx}
\usepackage{booktabs}
\usepackage{multicol}
\usepackage{multirow}
\usepackage{xcolor}
\usepackage{colortbl}
\usepackage{amsmath}
\usepackage{mathrsfs}
\usepackage{appendix}
\usepackage[accsupp]{axessibility}  
\newcommand{\greennumber}[1]{\textcolor[rgb]{0,1,0}{#1}}
\newcommand{\rednumber}[1]{\textcolor[rgb]{1,0,0}{#1}}

\usepackage[pagebackref,breaklinks,colorlinks,citecolor=eccvblue]{hyperref}

\usepackage{orcidlink}

\begin{document}

\title{Pensieve: Retrospect-then-Compare Mitigates Visual Hallucination} 


\author{
Dingchen Yang\inst{1} \and
Bowen Cao\inst{2} \and
Guang Chen\inst{1} \and
Changjun Jiang\inst{1}
}

\authorrunning{D.~Yang et al.}


\institute{
Tongji University \\
\email{\{dingchen\_yang,guangchen,cjjiang\}@tongji.edu.cn} \and
Peking University \\
\email{cbw2021@stu.pku.edu.cn}
}

\maketitle
\vspace{-0.4cm}

\begin{abstract}

    Multi-modal Large Language Models (MLLMs) demonstrate remarkable success across various vision-language tasks.
    However, they suffer from visual hallucination,
    where the generated responses
    diverge from the provided image.
    Are MLLMs 
    oblivious to the accurate visual cues when they hallucinate?
    Our investigation reveals that
    the visual branch
    may 
    equally
    advocate both accurate and 
    erroneous content.
    To address this issue,
    we propose
    \textit{Pensieve}\footnote{Code is available at \url{https://github.com/DingchenYang99/Pensieve}.},
    a training-free method
    that leverages the analogous visual hallucinations,
    which are induced by
    images sharing common semantic and appearance characteristics,
    to mitigate hallucination.
    Specifically,
    \textit{Pensieve} enables MLLMs to retrospect relevant images as references
    and compare 
    their visual content
    with the test image
    via confidence score subtraction.
    Moreover,
    our paradigm
    balances the effects of
    addressing errors from both the visual and textual branches by adaptively scaling the subtracted scores.
    Experiments on Whoops,
    LLaVA Bench,
    POPE, and MME
    demonstrate the efficacy of \textit{Pensieve} in mitigating visual hallucination,
    surpassing other advanced decoding strategies.
    \textit{Pensieve} also
    aids MLLMs
    in identifying visual details
    and
    enhance the specificity of generated image descriptions.
  
  \keywords{Visual Hallucination \and Multi-modal Large Language Model}
\end{abstract}

\section{Introduction}
\label{sec:intro}

Multi-modal Large Language Models (MLLMs)
have emerged as dominant forces in vision-language tasks~\cite{tong2024cambrian,zhang2024internlm,liu2023improvedllava,instructblip,zhu2023minigpt,chen2023shikra,lin2023sphinx,li2023blip,you2023ferret,yuan2023osprey,zhang2023gpt4roi}, 
showcasing remarkable advancements in comprehending a wide array of visual concepts.

\begin{figure}[tb]
  \centering
  \includegraphics[height=3.6cm]{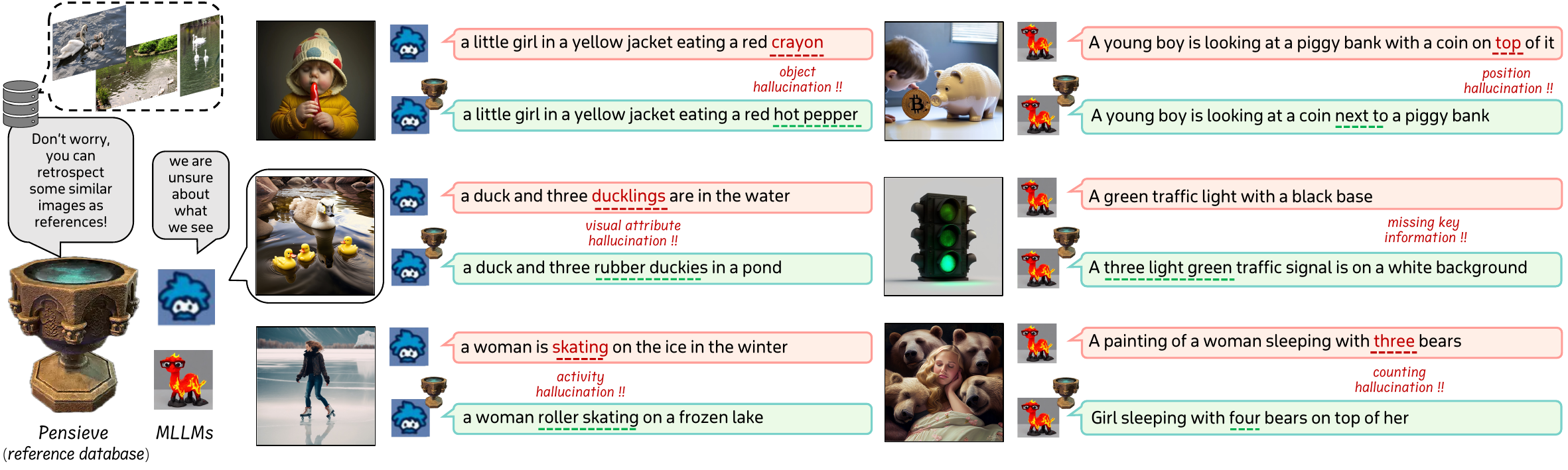}
  \caption{
    Visual hallucinations in image captions
    and an illustration of our proposed method \textit{Pensieve}.
    During inference, MLLMs
    are enabled to retrospect relevant images
    as references and compare their visual content with the test image.
    \textit{Pensieve}
    is capable of correcting erroneous
    object categories, attributes, activities, position, and numbers,
    wherever they occur in the sentence.
    \textit{Pensieve} also facilitates MLLMs to identify visual details in the image
    (\eg, the traffic light with three green lights).
  }
  \label{fig:caption_results}
\vspace{-.4cm}
\end{figure}

Despite their impressive capabilities,
state-of-the-art MLLMs are susceptible to visual hallucination~\cite{liu2024survey,chen2024alleviating,yan2024evaluating,huang2023opera,jiang2023hallucination,jing2023faithscore,leng2023mitigating,yin2023woodpecker,zhou2023analyzing,wang2023evaluation,tong2024eyes,chen2023mitigating,sun2023aligning,liu2023mitigating},
wherein they inaccurately describe the visual inputs.
Specifically, MLLMs can generate conflicting or fabricated content that
diverges from the provided image, and may overlook crucial visual details.
Examples illustrating this issue are presented in~\cref{fig:caption_results}.
Throughout this paper,
we refer to the generated tokens
(word or subword)
containing inaccurate semantics as hallucinatory tokens.

Understanding the genesis of visual hallucinations is paramount for hallucination reduction.
Previous studies underline flaws within MLLMs,
such as under-distinctive visual features~\cite{tong2024eyes},
the image-text modality gap~\cite{jiang2023hallucination,sun2023aligning},
biased feature aggregation patterns~\cite{wang2023evaluation,huang2023opera,lee2023volcano},
and fitting superficial language priors in the training data~\cite{zhou2023analyzing,leng2023mitigating}.
While these flaws impede MLLMs from accurately comprehending images,
our investigation suggests a different perspective,
that MLLMs may \textit{\textbf{not}} be 
completely unaware of
the accurate visual cues
when they hallucinate;
instead, their predictions reveal an uncertainty between 
accurate and hallucinatory content.
Specifically, we observe that 
MLLM's visual branch tends to equally advocate
both the accurate and 
erroneous
token candidates,
which we term \textit{visually deceptive candidates},
by contributing close confidence scores to them.
This ambiguity can lead to visual hallucination.
The most straightforward approach to
discern the accurate candidates from the visually deceptive ones
is directly adjusting the predicted confidence score distribution,
prioritizing the faithful content over hallucinations.
However, 
this objective is beyond the scope of existing distribution post-processors,
predominantly the Contrastive Decoding based methods~\cite{leng2023mitigating,favero2024multi,zhang2024debiasing},
which are adept at reducing the uni-modal bias embedded in the language decoder
yet omit errors from the visual branch.
Commence with a premise that 
similar images may
induce analogous visual hallucinations,
we step forward to
analyze the confidence score distribution shift
when replacing the test image with retrieved alternatives.
We observe moderate changes in visually deceptive candidates' scores,
whereas the scores for accurate candidates 
exhibit more significant variations.
Leveraging 
this phenomenon, 
we introduce a training-free approach \textit{Pensieve}.
During inference, MLLMs are enabled to retrospect pertinent images as references
and compare them with the test image.
Specifically,
the confidence scores
predicted by the test image and retrieved references
are 
subtracted
to promote
the candidates with sharp score variations,
which are \textit{\textbf{less}} likely to be the \textit{visually deceptive candidates}.
Moreover, \textit{Pensieve} 
retains the capability in debiasing erroneous language priors,
and balances its effect in addressing errors from the visual and textual branches
by adaptively scaling the subtracted scores.

We validate \textit{Pensieve}'s effectiveness in mitigating visual hallucination
on image captioning and Visual Question Answering (VQA) tasks.
Quantitative and qualitative results on four benchmarks,
Whoops~\cite{bitton2023breaking},
LLaVA Bench in the wild~\cite{liu2024visual},
MME~\cite{fu2023mme},
and POPE~\cite{li2023evaluating}
demonstrate the superiority of \textit{Pensieve},
notably increasing the FaithScore~\cite{jing2023faithscore} by 0.4
for LLaVA1.5~\cite{liu2023improvedllava} on Whoops,
and the total score by 55 for InstructBLIP~\cite{instructblip} on MME.
Additionally, 
\textit{Pensieve} helps MLLMs identify visual details
and generate more
specified image descriptions.

\textbf{Our main contributions are three-fold}:
\begin{itemize}
\item We empirically reveal that
MLLMs are sometimes \textbf{\textit{aware}} of accurate visual cues amidst visual hallucination,
and analogous visual hallucinations can occur among similar images,
which can be leveraged to 
mitigate hallucination.

\item we introduce \textit{Pensieve},
a novel paradigm that allows MLLMs to retrospect similar images as references,
and discern accurate visual cues through comparing confidence scores
predicted by the visual references.
Moreover, the subtracted scores are adaptively scaled to balance the effect
in mitigating hallucinations stemming from MLLMs' visual or textual branch.

\item Experiments on image captioning and VQA
demonstrate the superiority of \textit{Pensieve},
\eg, +0.4 FaithScore on Whoops for LLaVA1.5,
and +55 total scores on MME for InstructBLIP.
Qualitative results demonstrate that
\textit{Pensieve} also
enhances the specificity of image descriptions.
\end{itemize}

\section{Related works}

\subsection{Visual Hallucination and their Origins}

Visual hallucination~\cite{liu2024survey} refers to the issue 
wherein the descriptive content
diverges from the visual input.
These erroneous responses may exhibit fabrication, contradictions, or scarce specificity to the provided image.
Initial investigations primarily address object-level visual hallucination~\cite{rohrbach2018object,li2023evaluating,zhou2023analyzing},
focusing solely on inappropriate nouns.
This problem is subsequently extended to a finer granularity~\cite{fu2023mme,chen2023mitigating,wang2023amber,tong2024eyes,liu2024phd,zhang2024numberhallucination,wu2024evaluating},
including errors in visual attributes, spatial relationships, physical states, activities, and numbers.

The origins of visual hallucination stem from various sources.
Some highlight flaws in the visual encoder,
such as 
limited image resolution~\cite{zhai2023halle},
under-distinctive visual representations that
lack visual details~\cite{tong2024eyes} or spatial information~\cite{chen2023mitigating},
and poor cross-modal representation alignment~\cite{jiang2023hallucination}.
Others emphasize deficiencies within the language decoder,
such as biased attention score distribution~\cite{lee2023volcano,wang2023evaluation,huang2023opera,wu2024noiseboost},
adherence to superficial syntactical patterns
(\eg, frequent answers~\cite{liu2023mitigating},
contextual co-occurrence of objects~\cite{zhou2023analyzing,leng2023mitigating},
and activities~\cite{wang2024mementos}),
the overwhelming parametric knowledge~\cite{shi2023trusting,zhai2023halle},
and error snowballing~\cite{zhang2023language,zhou2023analyzing,yue2024lessismore}.
In this study, 
we investigate the genesis of visual hallucination,
and find that MLLM's
visual branch tends to equally advocate both the accurate and erroneous candidates.
This observation 
suggests new perspectives that visual hallucination can be reduced
\textit{\textbf{without}} directly tackling MLLMs' inherent deficiencies,
saving enormous amount of labeling or training costs.

\subsection{Mitigating Visual Hallucination}

\subsubsection{Parameter Tuning.}
methods include 
curating diverse multi-modal instruction tuning dataset~\cite{liu2023mitigating},
designing feedback systems~\cite{sun2023aligning, yu2024rlaif},
and the provision of extra supervisions~\cite{jiang2023hallucination,chen2023mitigating}.
Nonetheless, 
the training cost becomes prohibitive for large-scale MLLMs.
Furthermore, excessive parameter tuning may impair 
strengths of MLLMs~\cite{liu2023mitigating,chen2023mitigating},
when the training recipe is suboptimal.

\subsubsection{Model Ensemble.}
Integrating knowledge from other models compensates for MLLMs' shortcomings.
Feasible methods include improving the object recognition 
accuracy through ensembling detectors~\cite{chen2024halc,zhao2024mitigating},
and obtaining distinctive
features by ensembling multiple vision encoders~\cite{zhao2024mitigating}.
Another line of work 
utilizes
a language model to post-hoc revise
visual hallucinations~\cite{yin2023woodpecker,zhou2023analyzing}.
Key challenges within this paradigm include 
tailoring interfaces for various 
task-specific
models,
and automating their selection
based on the hallucination categories.

\subsubsection{Decoding Strategy.}
Adjusting the confidence score distribution straightforwardly
is a more efficient approach compared to model training and ensemble.
OPERA~\cite{huang2023opera} directly discards the candidates 
that may skew subsequent content toward hallucination and reelect the others.
VCD~\cite{leng2023mitigating} and its variants~\cite{favero2024multi,zhang2024debiasing,an2024agla}
extends Contrastive Decoding 
(CD)~\cite{zhao2021calibrate,li2022contrastive},
which aims to mitigate factual hallucinations in LLMs,
to the vision domain.
VCD distorts the visual input to amplify language priors,
and downgrades the candidates
advocated merely by the language priors
through logits subtraction.
Other
CD based methods~\cite{wang2024icd,chen2024hio}
also
effectively reduce the erroneous language bias embedded in the decoder of MLLMs.
However, hallucinations originating from the visual branch 
have remained unexplored.
In this study, we
investigate how the visual modality
mistakenly promotes erroneous candidates,
and propose a novel method \textit{Pensieve} to
mitigate hallucinations
that stem from MLLMs' visual or textual branch.

\section{Delve into Visual Hallucination}

In this section,
we investigate the genesis of 
visual hallucination
through an end-to-end
approach.
Commence with a fundamental question: 
to what extent are MLLMs 
unaware of accurate visual cues amidst hallucinations?
We designed a token-by-token evaluation pipeline 
and present our findings.

\subsubsection{Background.}

Leading MLLMs~\cite{liu2023improvedllava,instructblip,zhu2023minigpt,you2023ferret,chen2023shikra,lin2023sphinx}
incorporate auto-regressive language models~\cite{zheng2023judging,touvron2023llama},
which repeatedly selects the next token from their vocabulary $\mathcal V$ based on the probability of each token candidate $x_i$,
\begin{equation}
  p_\theta(x_i|\boldsymbol{v},\boldsymbol{x},\boldsymbol{y}_{<t}) = \frac{\exp{(\boldsymbol{h}_t \cdot E_c(x_i)})}{\sum_{x' \in \mathcal V} \exp{(\boldsymbol{h}_t \cdot E_c(x')})}
  \label{eq:probability}
\end{equation}
where $\boldsymbol{v}$ are the visual inputs,
$\boldsymbol{x}$ and $\boldsymbol{y}_{<t}$ are the prompt and past generated tokens, respectively.
$\boldsymbol{h}_t$ is the last hidden state predicted by the language model
(also known as the next token feature).
$E_c(x_i)$ is the token embedding of candidate $x_i$ in the language head.
$(\cdot)$ is the inner product operator.
The confidence score $\boldsymbol{h}_t \cdot E_c(x_i)$ 
embodies the degree of dominance of $x_i$'s semantics in $\boldsymbol{h}_t$\footnote{We find that for LLaVA1.5
and InstructBLIP,
the mean L2 norm of all $E_c(x')$, for $x'$ in $\mathcal V$, are close to 1. Thus the inner product result is close to the projection coordinate of the next token feature on a candidate's embedding in the language model head.}.

\subsubsection{Analysis Pipeline.}
\label{sect:motivation}
\begin{figure}[tb]
  \centering
  \includegraphics[height=6.1cm]{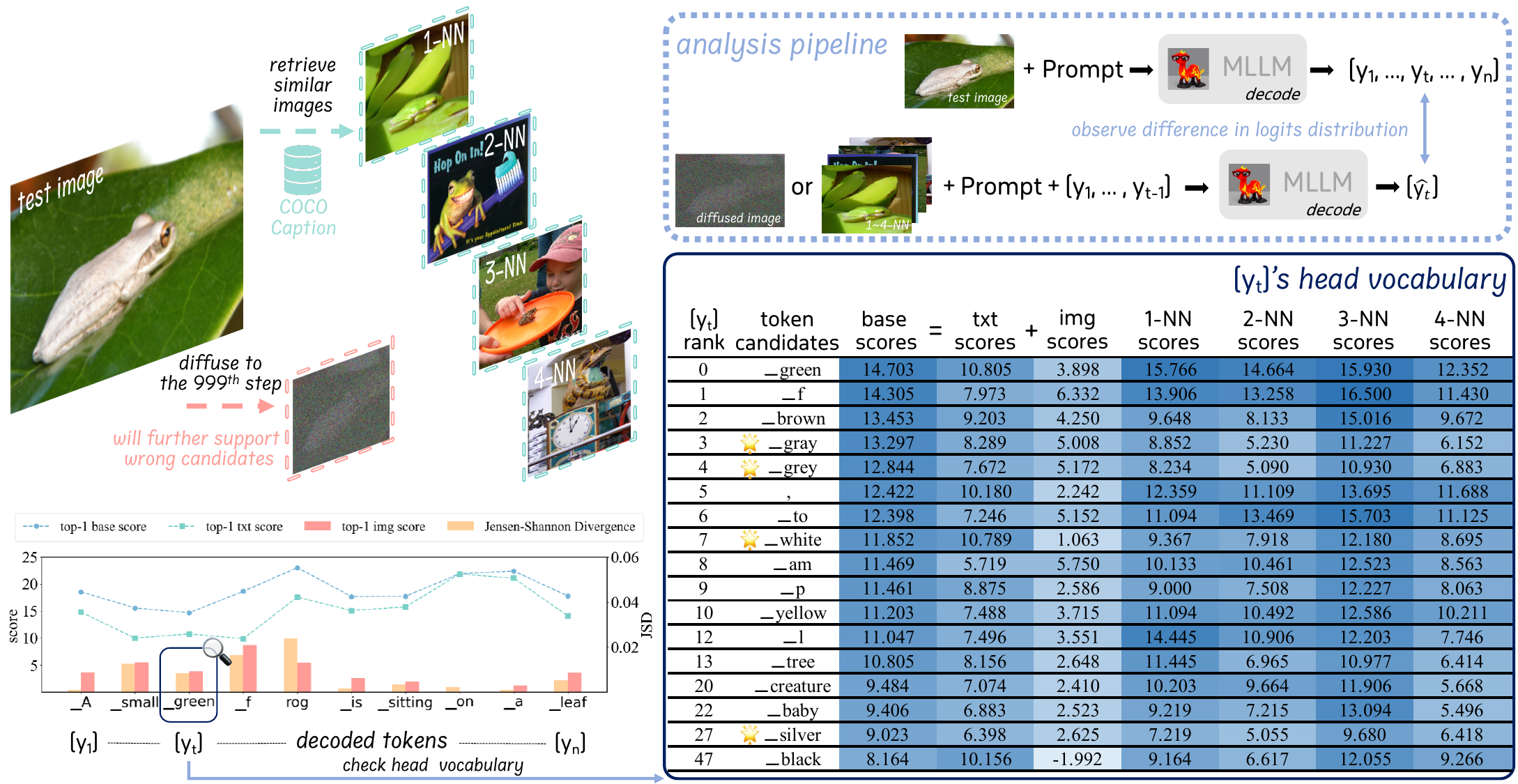}
  \caption{
    Our visual hallucination analysis pipeline and results.
    We investigate LLaVA1.5's
    predictions
    with alternative visual inputs in the same context.
    Such that the difference between $y_t$'s and $\hat{y_t}$'s confidence score distribution 
    manifests the influence of the \textbf{\textit{visual}} modality to MLLMs' prediction.
    We find that LLaVA1.5 is 
    aware of the accurate visual cues amidst hallucination,
    as the visual information contributed +5.008 scores for the accurate candidate $\_gray$.
    However, the visual input also erroneously advocates for inaccurate candidates
    $\_green$ (+3.898) and $\_brown$ (+4.250).
    We also observe that images with similar semantics and appearance can induce
    analogous visual hallucinations,
    and leverage this phenomenon to assist MLLMs in discerning accurate content.
    This test sample is from OpenImages validation set~\cite{kuznetsova2020open}.
  }
  \label{fig:gen_problem}
\vspace{-.35cm}
\end{figure}

In this study, 
we pose the following questions:
\textit{When visual hallucination occurs,
are MLLMs completely ignorant of the accurate visual cues?
If not,
can we help them distinguish the accurate content from hallucinations?}
To address these inquiries,
we aim to decouple the contribution of visual information in $\boldsymbol{v}$ to the predicted confidence scores.
Inspired by~\cite{lin2023unlocking}, 
we first input the test image $\boldsymbol{v}^{\tau}$
into the MLLM
and greedily decode tokens $[y_1, \dots, y_n]$.
At decoding step $t$,
We denote $y_t$'s confidence score
distribution as the \textit{base scores}.
Next, the test image is replaced with alternatives $\boldsymbol{v}'$ 
(either a noised
image without valid visual content,
or other similar images),
and the predicted tokens thus far $[y_1, \dots, y_{t-1}]$
are concatenated to $\boldsymbol{x}$
to predict a new hidden state $\hat{\boldsymbol{h}_t}$ and decode $\hat{y_t}$,
as illustrated in~\cref{fig:gen_problem}.
We anticipate the input shift vectors $\Delta \boldsymbol{ve} = VE(\boldsymbol{v}^{\tau}) - VE(\boldsymbol{v}')$, 
\ie, the change in input visual information,
to induce a corresponding output feature shift $\Delta \boldsymbol{h} = \boldsymbol{h}_t - \hat{\boldsymbol{h}_t}$. 
Formally,
\begin{equation}
    \boldsymbol{h}_t = LLM_{\theta}([VE(\boldsymbol{v}^{\tau});TE(\boldsymbol{x};\boldsymbol{y}_{<t})])
    \label{eq:LLM1}
\vspace{-0.4cm}
\end{equation}
\begin{equation}
    \hat{\boldsymbol{h}_t} = LLM_{\theta}([VE(\boldsymbol{v}');TE(\boldsymbol{x};\boldsymbol{y}_{<t})])
    \label{eq:LLM2}
\end{equation}
where $VE$ denotes 
the visual encoder and 
adaptor.
$TE$ is the text embedding layer.
Therefore, the confidence score distribution shift $\Delta \boldsymbol{h} \cdot E_c(x')$ for $x' \in \mathcal V$
(the subtraction of $\hat{y_t}$'s confidence scores from $y_t$'s)
represents the semantics in $\boldsymbol{h}_t$ contributed by the visual information in $\Delta \boldsymbol{ve}$.

\subsection{MLLMs are not Blind amidst Hallucination}
\label{sect:preexp}
To integrally decouple the visual branch's contribution to the prediction,
$\Delta \boldsymbol{ve}$ should encapsulate the \textbf{\textit{majority}} of visual information present in the image.
Following the diffusion process~\cite{ho2020denoising},
we render the test image nearly indistinguishable from Gaussian noise.
Then MLLM \textit{blindly} predicts a new score distribution
devoid of input visual cues
(the \textit{txt scores} 
in~\cref{fig:gen_problem}).
Consequently,
the subtraction of the blindly predicted \textit{txt scores} from the original \textit{base scores}
can be interpreted as the contribution of the visual modality
(the \textit{img scores}).

\cref{fig:gen_problem}
exhibits the multimodally distributed \textit{img scores} across
representative top-ranked candidates.
Notably,
both the accurate candidate (\eg, $\_gray$)
and other erroneous candidates (\eg, $\_green$, $\_brown$ and $\_yellow$)
have relatively high \textit{img scores},
indicating comparable advocacy from the visual branch.
Therefore, the MLLM~\cite{liu2023improvedllava} is \textit{\textbf{not}} completely ignorant\footnote{
We quantify the extent of dependency of MLLMs' predictions on the visual input with the
\textit{img scores} and the
Jensen-Shannon Divergence.
Details are in Appendix.
},
but uncertain,
to the accurate visual information when it hallucinates.
We designate these 
inaccurate
candidates,
which are mistakenly advocated by the visual input,
as \textit{visually deceptive candidates}.
Conversely,
candidates with limited support, even opposition, from the visual branch are termed \textit{textually deceptive candidates}
(\eg, candidate $\_black$ gets -1.992 \textit{img scores}).

\subsection{Visual References Help Discern Accurate Content}
\label{sect:effectofnns}
In the presence of both visually and textually deceptive candidates,
implementing VCD~\cite{leng2023mitigating},
which adds scaled \textit{img scores} to \textit{base scores},
will further elevate the scores of the visually deceptive candidates $\_green$ and $\_brown$,
thereby exacerbating visual hallucination.
In this case, it also risks depressing the specificity by further promoting candidate $\_f$ (first token of frog).
We continue to explore feasible methods 
to help MLLMs discern these
erroneous content.
Start with an intuitive hypothesis that in the same context $\boldsymbol{x}+\boldsymbol{y}_{<t}$,
images sharing similar semantics and appearance
are likely to induce \textbf{\textit{analogous}} visual hallucinations,
we proceed by analyzing the scores
predicted using retrieved images,
which possess visual characteristics in common with the test image.
We present four retrieved images from COCO-Caption~\cite{lin2014microsoft},
with corresponding predictions in the \textit{k-NN} \textit{scores} columns in~\cref{fig:gen_problem}.

Upon comparing the \textit{base scores} with four \textit{k-NN} \textit{scores},
a notable observation emerges: 
the score of the accurate candidate $\_gray$ decreases significantly
from 13.297 to 7.865 in average across four references.
In contrast,
the score change of
visually deceptive 
(\eg, $\_green$, $\_brown$, and $\_yellow$)
and textually deceptive (\eg, $\_black$)
candidates are relatively modest
(\eg, candidate $\_green$ scores 15.390 and 12.352 in images without green frogs,
and candidate $\_brown$ also has relatively high scores in images without brown frogs).
This observation suggests that
in the same context, similar images can induce analogous visual hallucinations.
Furthermore,
token candidates whose scores vary significantly
across similar images in the same context are \textit{more likely to be the correct ones}
than candidates with moderate changes.
We leverage 
this phenomenon to
discern accurate content and mitigate visual hallucination.

\section{Methodology}
Motivated by our observation that
analogous hallucinations among similar images
can help MLLMs discern accurate content,
we propose a training-free and plug-and-play method \textit{Pensieve},
which
comprises two key components,
searching similar images and contrasting visual cues.

\begin{figure}[tb]
  \centering
  \includegraphics[height=4.3cm]{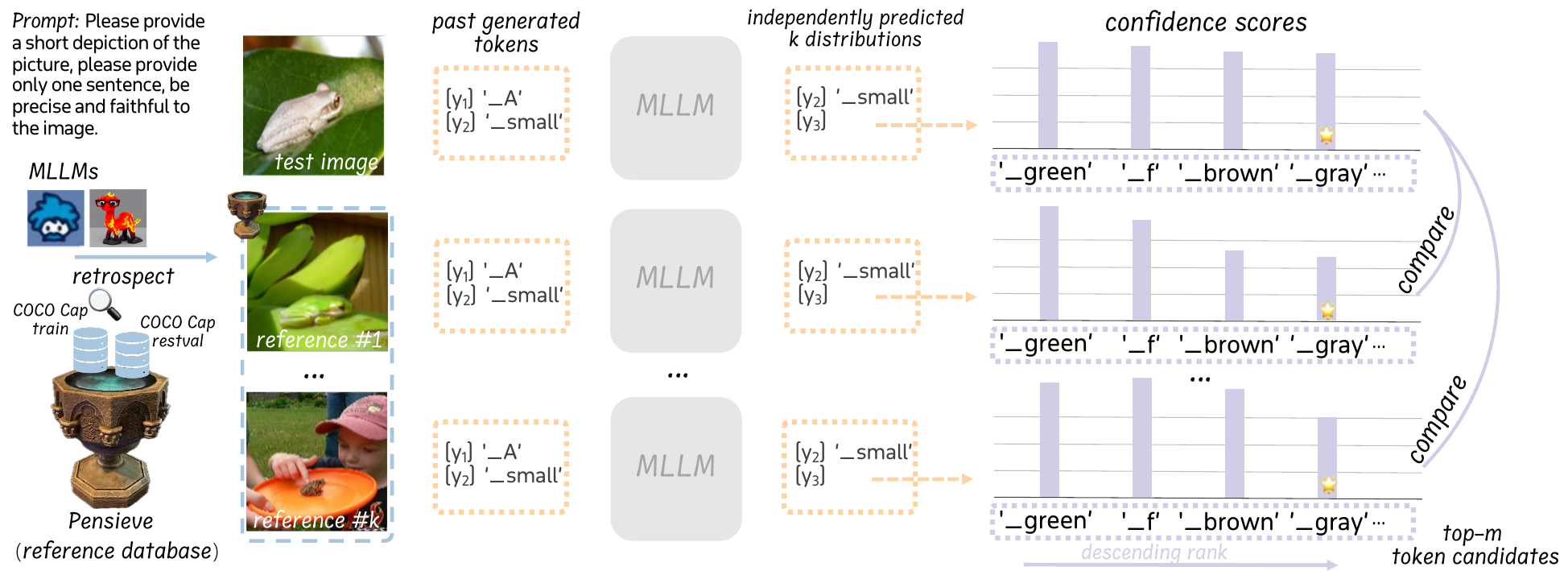}
  \caption{
    Our approach identifies erroneous candidates
    that are mistakenly supported by the visual branch
    by leveraging the analogous visual hallucinations among similar images.
    Our reference database comprises a variety of images.
    During inference,
    relevant references are retrieved from this database,
    and MLLM generates distinct prediction for each reference
    in the same context.
    The predicted scores are then subtracted to highlight the accurate candidates.
  }
  \label{fig:method}
\vspace{-.45cm}
\end{figure}

\subsection{Retrospect Visual Concepts}
\label{sect:retrospect}
We aim to build a reference database
that encompasses
diverse visual concepts for MLLMs to retrospect.
Concretely, our database contains samples from
the COCO Caption~\cite{lin2014microsoft} dataset,
which covers
a broad spectrum of
daily and common visual content.
Moreover,
the visual references
should be capable of
\textit{\textbf{inducing}} visual hallucinations,
such that the analogous hallucinations can be further leveraged to reduce hallucination.
Therefore,
we include both
the Karpathy~\cite{karpathy2015deep} \textit{train} and \textit{restval} splits
of COCO Caption.
We posit that images from the \textit{restval} split are more likely to induce
visual hallucinations,
as they are \textit{\textbf{not}} included in the MLLMs' training set~\cite{liu2023improvedllava,instructblip},
and consequently benefit model performance.
Specifically, 
we retrieves $k$ most relevant visual references 
for each test sample $q$
from the reference database $\mathcal D$,
based on a similarity measure $\mathcal F (\cdot,\cdot)$. Formally,
\vspace{-0.1cm}
\begin{equation}
    \{r_{1}, \dots, r_{k} | q\} = \underset{k}{\operatorname{arg\,max}} \{ \mathcal F ( E_{R}(s_{j}), E_{R}(q)) | s_{j} \in \mathcal D \}
    \label{eq:retrieve}
\vspace{-0.1cm}
\end{equation}
where $E_{R}(\cdot)$ is the retriever that embeds the raw inputs into vector representations.
$\mathcal D[r], r \in \{r_{1}, \dots, r_{k} | q\}$, are the desired references for $q$.
For image captioning,
the visual feature $E_{R}(\boldsymbol{v})$ is used to retrieve images with similar semantics and appearance.
In binary VQA task,
we adapt our approach to find visual references
that align with the semantic of the question using $E_{R}(\boldsymbol{x})$.
We expect that comparing with such references
will aid in reducing false positives
(A detailed case study and discussion are in~\cref{sec:case_study_mme}). 

\subsection{Compare Visual Concepts}

Contrasting with the confidence scores
(denoted as \textit{logits} in \cref{eq:contrast} and \cref{eq:adaptive3})
predicted by visual references
can help distinguish accurate visual cues.
As illustrated in \cref{fig:method},
the MLLM generates $k+2$ distinct predictions for each token candidate $x_i \in \mathcal V$,
\ie, from one test image $\boldsymbol{v}^{\tau}$,
one diffused image $\boldsymbol{v}^{d}$,
and $k$ retrieved images $\{\boldsymbol{v}^{NN}\}_{k}$,
with identical textual prefix $\boldsymbol{x}+\boldsymbol{y}_{<t}$.
Note that the context length is \textbf{\textit{not}} increased in our paradigm,
such that the computational burden is not significantly increased.
We contrast $x_i$'s score corresponding to the test image $\boldsymbol{v}^{\tau}$ to the $k+1$ references.
Formally,
\begin{align}
    & \mathit{logits}(x_i|\boldsymbol{x},\boldsymbol{y}_{<t},\boldsymbol{v}^{\tau},\boldsymbol{v}^{d},\{\boldsymbol{v}^{NN}\}_{k}) \notag \\
    & =(\alpha_{\tau}+\alpha_{d}^{t}+\alpha_{NN}^{t})\mathit{logits}(x_i|\boldsymbol{x},\boldsymbol{y}_{<t},\boldsymbol{v}^{\tau}) \notag \\
    & -\frac{\alpha_{NN}^{t}}{k}\sum_{j=1}^{k}\mathit{logits}(x_i|\boldsymbol{x},\boldsymbol{y}_{<t},\boldsymbol{v}^{NN}_j) - \alpha_{d}^{t}\mathit{logits}(x_i|\boldsymbol{x},\boldsymbol{y}_{<t},\boldsymbol{v}^{d})
    \label{eq:contrast}
\end{align}
The subtraction operator in Equation~\ref{eq:contrast} underscores the
difference between the test image and the visual references,
thereby downgrading the analogous visual hallucinations across similar images. 
The retrieved images $\{\boldsymbol{v}^{NN}\}_{k}$ target at reducing visually deceptive content,
and $\boldsymbol{v}^{d}$ aids in
addressing textually deceptive hallucinations.
The outcome of~\cref{eq:contrast}
is subsequently used for token selection,
and this process 
is activated
throughout the output sentence.

\subsubsection{Adaptive Logits Processing.}
We 
regulate the influence of 
$\{\boldsymbol{v}^{NN}\}_{k}$ and
$\boldsymbol{v}^{d}$
to balance \textit{Pensieve}'s effect in addressing errors from the
visual and textual branches.
Specifically, in cases where MLLMs exhibit uncertainty~\cite{zhou2023analyzing} 
regarding the recognized visual cues,
\ie, the scores $\{l^{\delta}_{i,t}\}_{i=0}^{m-1}$ calculated by~\cref{eq:adaptive3} 
demonstrate multimodal distribution,
then the coefficient $\alpha_{d}^{t}$ 
is reduced while
$\alpha_{NN}^{t}$
is increased
at the current decoding step $t$.
This adjustment ensures that
the visually deceptive candidates are not further endorsed.
Formally,
\vspace{-0.1cm}
\begin{equation}
  \alpha_{d}^{t} = \beta_{d}\exp{(\mathop{\max}(\mathrm{softmax}(\{l^{\delta}_{i,t}\}_{i=0}^{m-1})))}
  \label{eq:adaptive1}
\end{equation}
\vspace{-0.23cm}
\begin{equation}
  \alpha_{NN}^{t} = \beta_{NN}\exp{(1-\mathop{\max}(\mathrm{softmax}(\{l^{\delta}_{i,t}\}_{i=0}^{m-1})))}
  \label{eq:adaptive2}
\end{equation}
\vspace{-0.2cm}
\begin{equation}
    l^{\delta}_{i,t} = \mathit{logits}(x_i|\boldsymbol{x},\boldsymbol{y}_{<t},\boldsymbol{v}^{\tau}) -\mathit{logits}(x_i|\boldsymbol{x},\boldsymbol{y}_{<t},\boldsymbol{v}^{d}) \quad \text{s.t.} \quad x_i \in \mathcal V_{head}^{m}
    \label{eq:adaptive3}
\end{equation}
Following the adaptive plausibility constraint~\cite{li2022contrastive},
we only consider $x_i$ that are among the top-ranked $m$ candidates,
\ie, a fixed-length head vocabulary $\mathcal V_{head}^{m}$,
which is selected based on the scores predicted from $\boldsymbol{v}^{\tau}$.
We set $m$ to 50 for image captioning and 2 for binary VQA.
$\alpha_{\tau}$, $\beta_{d}$ and $\beta_{NN}$ are hyper-parameters,
which are by default set to 1.0, 0.1, and 0.1, respectively.

\section{Experiments}

\subsubsection{General Settings.}
We evaluate \textit{Pensieve} on
image captioning benchmarks, including
Whoops~\cite{bitton2023breaking}
and LLaVA-Bench~\cite{liu2024visual},
as well as on binary VQA benchmarks
(MME~\cite{fu2023mme}
and POPE~\cite{li2023evaluating}),
evaluating hallucinations in both generative and discriminative tasks.
The effectiveness of each component
is examined via ablation study.
All experiments are in zero-shot manner,
with LLaVA1.5-7B~\cite{liu2023improvedllava} and InstructBLIP-7B~\cite{instructblip} as baseline models,
both using Vicuna~\cite{zheng2023judging} as the language decoder\footnote{\url{https://github.com/lm-sys/FastChat}}.
We compare \textit{Pensieve} with two advanced decoding strategies VCD~\cite{leng2023mitigating}
and DoLa~\cite{chuang2023dola}.
We employ greedy decoding as the baseline strategy for reproducibility,
and to avoid visual hallucinations induced by token sampling~\cite{lee2022factuality,li2024dawn,wang2023evaluation}.
Implementation details and specific experimental settings for each benchmark are in the Appendix.


\subsection{Image Captioning}
In light of that
MLLMs encode rich real-world conventions
and may embed certain superficial 
syntactical patterns
in model parameters~\cite{zhou2023analyzing},
we assume that commonsense-violating images 
(see examples in~\cref{fig:caption_results})
can highlight the problem of visual hallucination.
For quantitative evaluation, 
we employ image captioning metrics
BLEU~\cite{papineni2002bleu}, METEOR~\cite{denkowski2014meteor}, CIDEr~\cite{vedantam2015cider}, and SPICE~\cite{anderson2016spice}.
FaithScore~\cite{jing2023faithscore} is utilized for automatic visual hallucination evaluation,
which is calculated by ChatGPT
and a visual entailment expert model OFA\footnote{\url{https://github.com/OFA-Sys/OFA}}~\cite{wang2022ofa}.
We also provide qualitative results
on LLaVA Bench in the Wild~\cite{liu2024visual} 
in~\cref{fig:llavabench} and in the Appendix
to illustrate the efficacy of \textit{Pensieve} in
various image domains.

\subsubsection{Quantitative Results on Whoops.}
\begin{table}[tb]
\renewcommand\arraystretch{1.02}
  \caption{
    Results on Whoops benchmark.
    \textit{Pensieve} boosts the overall performance for 
    LLaVA-1.5
    and InstructBLIP.
    B4, M, C, S, and FS\% refer to Bleu@4, METEOR, CIDEr, SPICE,
    and FaithScore,
    respectively.
    We copy the results of 
    EVCap
    with and without retrieval augmentation.
  }
  \label{tab:whoops}
  \centering
  \scalebox{0.8}{
  \begin{tabular}{p{3.9cm}p{1.3cm}p{1.1cm}p{1.1cm}p{1.1cm}p{1.1cm}p{1.1cm}}
    \toprule
    \multicolumn{2}{c}{Method} & B4 $\uparrow$ & M $\uparrow$ & C $\uparrow$ & S $\uparrow$ & FS\% $\uparrow$\\
    \midrule
    \rowcolor{gray!20}
    \multicolumn{7}{c}{\textbf{\textit{further trained on COCO}}} \\
    \multicolumn{2}{l}{EVCap~\cite{li2023evcap} \scalebox{0.8}{\color{gray}{Vicuna-13B}}} & 24.1 & 26.1 & 85.3 & 17.7 & - \\
    \multicolumn{2}{l}{EVCap w/ RA~\cite{li2023evcap} \scalebox{0.8}{\color{gray}{Vicuna-13B}}} & 24.4 & 26.1 & 86.3 & 17.8 & - \\
    \midrule
    \rowcolor{gray!20}
    \multicolumn{7}{c}{\textbf{\textit{zeroshot}}} \\
    \cline{3-7}
    \multirow{4}*{LLaVA-1.5~\cite{liu2023improvedllava} \scalebox{0.8}{\color{gray}{Vicuna-7B}}} & greedy & 19.7 & 25.6 & 67.9 & 17.3 & 67.9 \\
    ~ & +{\it DoLa} & 19.9 & 25.6 & 67.8 & 17.4 & 67.7 \\
    ~ & +{\it VCD} & 19.1 & 25.4 & 69.1 & 17.3 & 67.6 \\
    ~ & +{\it Ours} & \bf 20.0 & \bf 26.3 & \bf 75.5 & \bf 17.8 & \bf 68.3 \\
    \cline{3-7}
    \multirow{4}*{InstructBLIP~\cite{instructblip} \scalebox{0.8}{\color{gray}{Vicuna-7B}}} & greedy & 24.9 & 26.5 & 87.3 & 18.2 & 74.1 \\
    ~ & +{\it DoLa} & 24.8 & 26.5 & 87.4 & 18.2 & \bf 74.2 \\
    ~ & +{\it VCD} & 25.5 & 27.0 & 89.2 & 18.2 & 72.2 \\
    ~ & +{\it Ours} & \bf 25.7 & \bf 27.1 & \bf 90.6 & \bf 18.6 & 74.0 \\
  \bottomrule
  \end{tabular}}
\end{table}

\cref{tab:whoops} demonstrate that our proposed \textit{Pensieve}
contribute to a substantial improvement in all captioning metrics
for both MLLMs,
notably increasing the CIDEr score by 7.6 and the Faithscore by 0.4 for LLaVA1.5,
surpassing DoLa and VCD in varying degrees.
Furthermore,
our retrospect-then-compare paradigm
exhibits more pronounced improvements 
compared to the traditional
retrieval-augmented (RA) image captioning paradigm EVCap~\cite{li2023evcap},
which relies on retrieved visual semantics to augment the input visual embeddings,
and increases the CIDEr score by only 1.0.

\subsubsection{Qualitative Results on LLaVA Bench.}
\begin{figure}[tb]
  \centering
  \includegraphics[height=4.3cm]{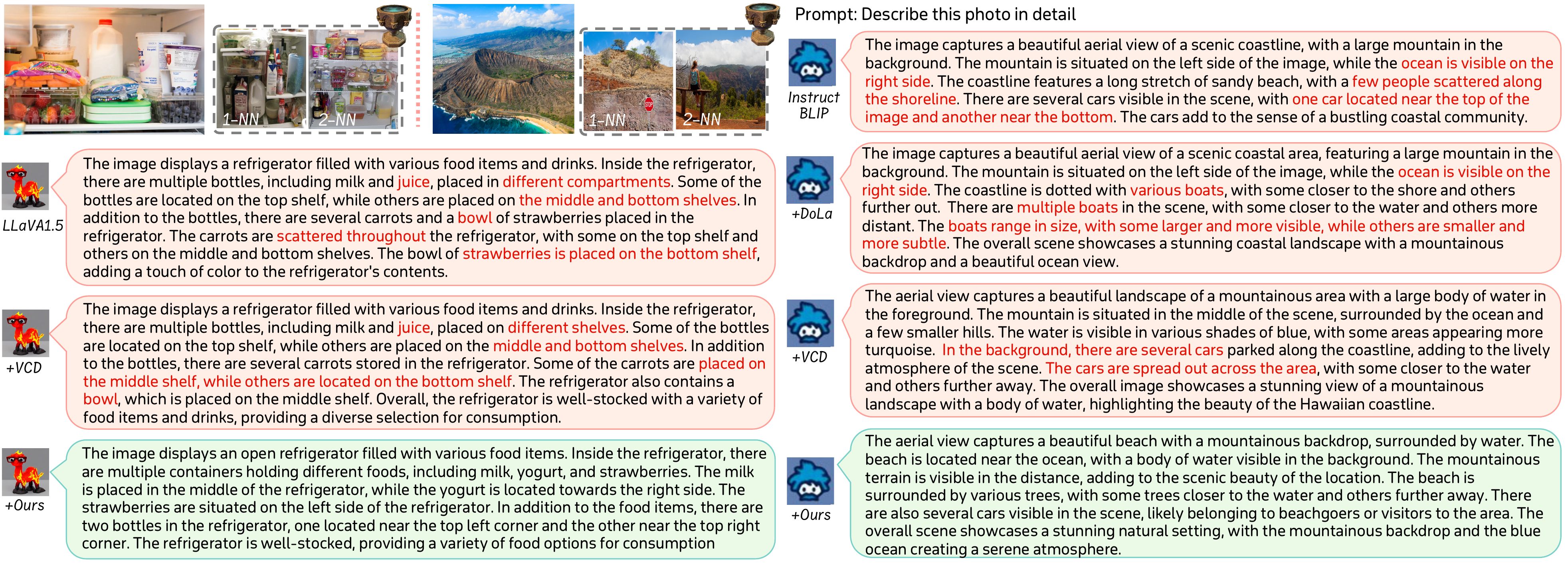}
  \caption{
    Qualitative results on LLaVA-Bench in the wild.
    \textit{Pensieve} effectively mitigates visual hallucination for
    both MLLMs,
    while VCD and DoLa
    may induce extra hallucinations.
    We present the test images with corresponding visual references,
    which illustrate a similar scenario but exhibit nuanced differences compared to the test image.
    We omit DoLa's result for LLaVA1.5 as it is identical to the original one.
  }
  \label{fig:llavabench}
\vspace{-.35cm}
\end{figure}

LLaVA Bench in the Wild
evaluates the ability of MLLMs to comprehend diverse visual concepts,
spanning indoor and outdoor scenes.
~\cref{fig:llavabench} demonstrates that
\textit{Pensieve} effectively reduces visual hallucinations at varying granularity
for both MLLMs,
\eg, it helps LLaVA1.5 correctly describe the spatial relationship between the milk, yogurt, and strawberries
in the same refrigerator compartment;
and help InstructBLIP identify the cars near the beach.
In these complex scenarios,
VCD and DoLa
struggle to assist MLLMs in discerning non-existent content,
and may induce additional errors,
\eg, the cars and the boats.
More examples are in the Appendix.

\subsection{Visual Question Answering}
We further evaluate \textit{Pensieve} on binary VQA benchmarks,
which are designed to gauge the extent of visual hallucination in a yes-or-no discriminative manner.
To prevent information leakage,
we exclude all images that appeared in MME
and POPE
from the reference database,
such that the test samples themselves do not serve as references.

\subsubsection{Results on MME.}
Following previous works~\cite{yin2023woodpecker,leng2023mitigating},
we present results on the hallucination subset of MME in~\cref{tab:mme},
including Color, Count, Existence, and Position subtasks.
We report the officially defined metric that combines accuracy and accuracy+.
\textit{Pensieve} improves the total score
for both LLaVA1.5 (+10) and InstructBLIP (+55),
surpassing DoLa and VCD in average by 33.4 and 33.3 scores, respectively.
In particular,
\textit{Pensieve} effectively mitigates visual hallucination in color-related perception task for both MLLMs
(+10 scores for LLaVA1.5 and noticeably +33.3 scores for InstructBLIP).
\begin{table}[h]
\renewcommand\arraystretch{1.01}
  \caption{
    Results on the hallucination subset of the MME benchmark
    (Color, Count, Existence, and Position subtasks).
    \textit{Pensieve} elevates the total score for both MLLMs.
    Results of all perception subtasks are in the Appendix.
  }
  \label{tab:mme}
  \centering
  \scalebox{0.85}{
  \begin{tabular}{p{2.2cm}p{1.6cm}p{1.3cm}p{1.4cm}p{1.3cm}p{1.3cm}p{1.4cm}}
    \toprule
    Model & Decoding & Color $\uparrow$ & Count $\uparrow$ & Exist. $\uparrow$ & Pos. $\uparrow$ & Total $\uparrow$\\
    \midrule
    \multirow{4}*{LLaVA-1.5} & greedy & 155.0 & \bf 158.3 & 195.0 & 123.3 & 631.7 \\
    ~ & +{\it DoLa} & 153.3 & 158.3 & 195.0 & 123.3 & 630.0 \\
    ~ & +{\it VCD} & 148.3 & 158.3 & 190.0 & 126.7 & 623.3 \\
    ~ & +{\it Ours} & \bf 165.0 & 153.3 & \bf 195.0 & \bf 128.3 & \bf 641.7 \\
    \cline{2-7}
    \multirow{4}*{InstructBLIP} & greedy & 120.0 & 60.0 & \bf 185.0 & 50.0 & 415.0 \\
    ~ & +{\it DoLa} & 120.0 & 60.0 & 185.0 & 50.0 & 415.0 \\
    ~ & +{\it VCD} & 123.3 & 60.0 & 185.0 & 53.3 & 421.7 \\
    ~ & +{\it Ours} & \bf 153.3 & \bf 78.3 & 180.0 &  \bf 58.3 & \bf 470.0 \\
  \bottomrule
  \end{tabular}}
\vspace{-.15cm}
\end{table}

\subsubsection{Results on POPE.}
We present the averaged Accuracy and F1 score across the random, popular, and adversarial splits in~\cref{tab:pope}.
On three subsets of POPE,
\textit{Pensieve} boosts the overall Accuracy of both LLaVA1.5 (+0.25 in avg.)
and InstructBLIP (+0.91 in avg.),
outperforming VCD by 1.85 for LLaVA1.5 and by 1.12 for InstructBLIP, respectively.
It is worth noticing that greedy decoding avoids errors stemming from token sampling,
especially when the MLLMs' predictions have high perplexity,
thus significantly surpassing the multi-nominal sampling strategy reported in VCD.

\begin{table}[tb]
  \centering
  \caption{
    Results on the POPE benchmark.
    \textit{Pensieve} demonstrates superior performance to VCD and DoLa
    on MSCOCO, AOKVQA and GQA subsets.
  }
  \label{tab:pope}
  \scalebox{0.83}{
  \begin{tabular}{p{2.0cm}p{1.6cm}p{1.2cm}p{1.2cm}p{1.2cm}p{1.2cm}p{1.2cm}p{1.2cm}}
    \toprule
    \multirow{2}*{Model} & \multirow{2}*{Decoding} & \multicolumn{2}{c}{\cellcolor{gray!40}COCO} & \multicolumn{2}{c}{\cellcolor{gray!20}AOKVQA} & \multicolumn{2}{c}{\cellcolor{gray!40}GQA}\\
    ~ & ~ & {\bf Acc. $\uparrow$} & {\bf F1 $\uparrow$} & {\bf Acc. $\uparrow$} & {\bf F1 $\uparrow$} & {\bf Acc. $\uparrow$} & {\bf F1 $\uparrow$}\\
    \midrule
    \multirow{4}*{LLaVA-1.5} & greedy & 85.56 & 84.11 & 84.32 & 84.40 & 84.73 & 84.84 \\
    ~ & +{\it DoLa} & 85.38 & 83.86 & 84.30 & 84.33 & 84.71 & 84.76 \\
    ~ & +{\it VCD} & 85.59 & \bf 85.53 & 82.07 & 83.39 & 82.17 & 82.87 \\
    ~ & +{\it Ours} & \bf 85.80 & 84.58 & \bf 84.56 & \bf 84.91 & \bf 85.01 & \bf 85.38 \\
    \cline{2-8}
    \multirow{4}*{InstructBLIP} & greedy & 85.14 & 84.45 & 81.73 & 83.18 & 80.58 & 82.03 \\
    ~ & +{\it DoLa} & \bf 85.21 & \bf 84.54 & 81.74 & 83.22 & 80.56 & 82.03 \\
    ~ & +{\it VCD} & 84.42 & 83.62 & 81.50 & 82.78 & 80.90 & 82.05 \\
    ~ & +{\it Ours} & 85.12 & 84.37 & \bf 83.30 & \bf 83.88 & \bf 81.76 & \bf 82.15 \\
  \bottomrule
  \end{tabular}}
\end{table}


\subsection{Ablation Studies}
\label{sec:ablation}
We validate the effectiveness of each component in \textit{Pensieve}
by individually ablating them,
and evaluating the performance variation on image
captioning
task
(the Whoops benchmark).
The impact of the reference database is investigated by
adding or removing the data samples.
We also conduct a case study on how \textit{Pensieve} improves performance on binary VQA task in~\cref{sec:case_study_mme}.
\begin{table}[h]
\renewcommand\arraystretch{1.1}
  \caption{
    Ablation study on the Whoops benchmark. 
    Each component in \textit{Pensieve} has a positive impact on the overall performance.
    All experiments 
    use LLaVA-v1.5-7B as the MLLM (Exp.1).
    Results of \textit{Pensieve} are in Exp.2.
    The performance gap to Exp.2 are presented for Exp.3 to 9,
    which are marked in \greennumber{green} or \rednumber{red}.
  }
  \label{tab:whoops_ablation}
  \centering
  \scalebox{0.81}{
  \begin{tabular}{p{0.9cm}p{1.0cm}p{1.0cm}p{1.0cm}p{1.0cm}p{1.0cm}p{1.3cm}p{1.3cm}p{1.3cm}p{1.3cm}p{1.3cm}}
    \toprule
    \centering Exp. & \centering img. ret. & \centering diffuse img. & \centering comp. & \centering adapt. & \centering num. refs & B4 $\uparrow$ & M $\uparrow$ & C $\uparrow$ & S $\uparrow$ & FS\% $\uparrow$ \\
    \midrule
    \centering 1 & & & & & \centering n/a & 19.7 & 25.6 & 67.9 & 17.3 & 67.9 \\
    \rowcolor{gray!20}
    \centering 2 & \centering \checkmark & \centering \checkmark & \centering \checkmark & \centering \checkmark & \centering 4 & 20.0 & 26.3 & 75.5 & 17.8 & 68.3\\
    \centering 3 & \centering rand. & \centering \checkmark & \centering \checkmark & \centering \checkmark & \centering 4  & $18.7^{\scalebox{0.6}{$\pm$0.3}}_{\scalebox{0.7}{(\greennumber{-1.3})}}$ & $26.0^{\scalebox{0.6}{$\pm$0.1}}_{\scalebox{0.7}{(\greennumber{-0.3})}}$ & $72.4^{\scalebox{0.6}{$\pm$0.7}}_{\scalebox{0.7}{(\greennumber{-3.1})}}$ & $17.6^{\scalebox{0.6}{$\pm$0.1}}_{\scalebox{0.7}{(\greennumber{-0.2})}}$ & $67.1^{\scalebox{0.6}{$\pm$0.3}}_{\scalebox{0.7}{(\greennumber{-1.2})}}$ \\
    \centering 4 & \centering \checkmark & & \centering \checkmark & \centering \checkmark & \centering 4 & 20.1\scalebox{0.7}{(\rednumber{+0.1})} & 26.2\scalebox{0.7}{(\greennumber{-0.1})} & 73.2\scalebox{0.7}{(\greennumber{-2.3})} & 17.9\scalebox{0.7}{(\rednumber{+0.1})} & 69.0\scalebox{0.7}{(\rednumber{+0.7})} \\
    \centering 5 & \centering \checkmark & \centering \checkmark & \centering add. & \centering \checkmark & \centering 4 & 17.6\scalebox{0.7}{(\greennumber{-2.4})} & 23.4\scalebox{0.7}{(\greennumber{-2.9})} & 58.7\scalebox{0.7}{(\greennumber{-16.8})} & 16.3\scalebox{0.7}{(\greennumber{-1.5})} & 66.9\scalebox{0.7}{(\greennumber{-1.4})} \\
    \centering 6 & \centering \checkmark & \centering \checkmark & \centering \checkmark & & \centering 4 & 19.9\scalebox{0.7}{(\greennumber{-0.1})} & 26.3\scalebox{0.7}{(-0.0)} & 71.9\scalebox{0.7}{(\greennumber{-3.6})} & 17.7\scalebox{0.7}{(\greennumber{-0.1})} & 68.9\scalebox{0.7}{(\rednumber{+0.6})} \\
    \centering 7 & \centering \checkmark & \centering \checkmark & \centering \checkmark & \centering \checkmark & \centering 1 & 19.4\scalebox{0.7}{(\greennumber{-0.6})} & 26.3\scalebox{0.7}{(-0.0)} & 73.8\scalebox{0.7}{(\greennumber{-1.7})} & 17.5\scalebox{0.7}{(\greennumber{-0.3})} & 66.0\scalebox{0.7}{(\greennumber{-2.3})} \\
    \centering 8 & \centering \checkmark & \centering \checkmark & \centering \checkmark & \centering \checkmark & \centering 2 & 19.8\scalebox{0.7}{(\greennumber{-0.2})} & 26.3\scalebox{0.7}{(-0.0)} & 74.6\scalebox{0.7}{(\greennumber{-0.9})} & 17.7\scalebox{0.7}{(\greennumber{-0.1})} & 67.5\scalebox{0.7}{(\greennumber{-0.8})} \\
    \centering 9 & \centering \checkmark & \centering \checkmark & \centering \checkmark & \centering \checkmark & \centering 8 & 20.2\scalebox{0.7}{(\rednumber{+0.2})} & 26.4\scalebox{0.7}{(\rednumber{+0.1})} & 76.5\scalebox{0.7}{(\rednumber{+1.0})} & 17.9\scalebox{0.7}{(\rednumber{+0.1})} & 67.6\scalebox{0.7}{(\greennumber{-0.7})} \\
  \bottomrule
  \end{tabular}}
\vspace{-.35cm}
\end{table}

\subsubsection{Similar Images are Better than Random.}
We first ablate the image retrieval process (abbreviated img. ret.),
randomly sampling four images from the reference database instead.
The averaged results and the standard deviations of five separate runs
are shown in Exp.3.
The results suggest that random images can have positive effects
(+4.5 CIDEr score and +0.3 SPICE score compared to Exp.1),
highlighting the robustness of \textit{Pensieve} in 
scenarios where similar images are absent.
We present the random and retrieved images in~\cref{fig:ablation} 
to illustrate the difference
in similarity between visual references.
Moreover, when similar images are available (Exp.2),
\textit{Pensieve} contribute to a more significant performance gain across all metrics 
(CIDEr +7.6 and SPICE +0.5).
Note that we retain the diffused image as it helps debias certain erroneous language priors,
and the CIDEr score dropped by 2.3 in Exp.4 after the diffused image is ablated.

\subsubsection{Visual Concepts Comparison is Better than Combination.}
We fix hyper-parameters and alter the visual concept comparison process (abbreviated comp.),
transitioning from logits subtraction to addition in~\cref{eq:contrast},
\ie, emphasizing the common visual cues among various images.
Exp.5 demonstrates notable decreases in all metrics
(\eg, CIDEr -16.8),
indicating that the additive paradigm can
harm
the language modeling process.
This result validates that contrasting visual concepts is better than combining them.

\subsubsection{The Size and Content of the Reference Database.}
\begin{figure}[tb]
  \centering
  \includegraphics[height=4.5cm]{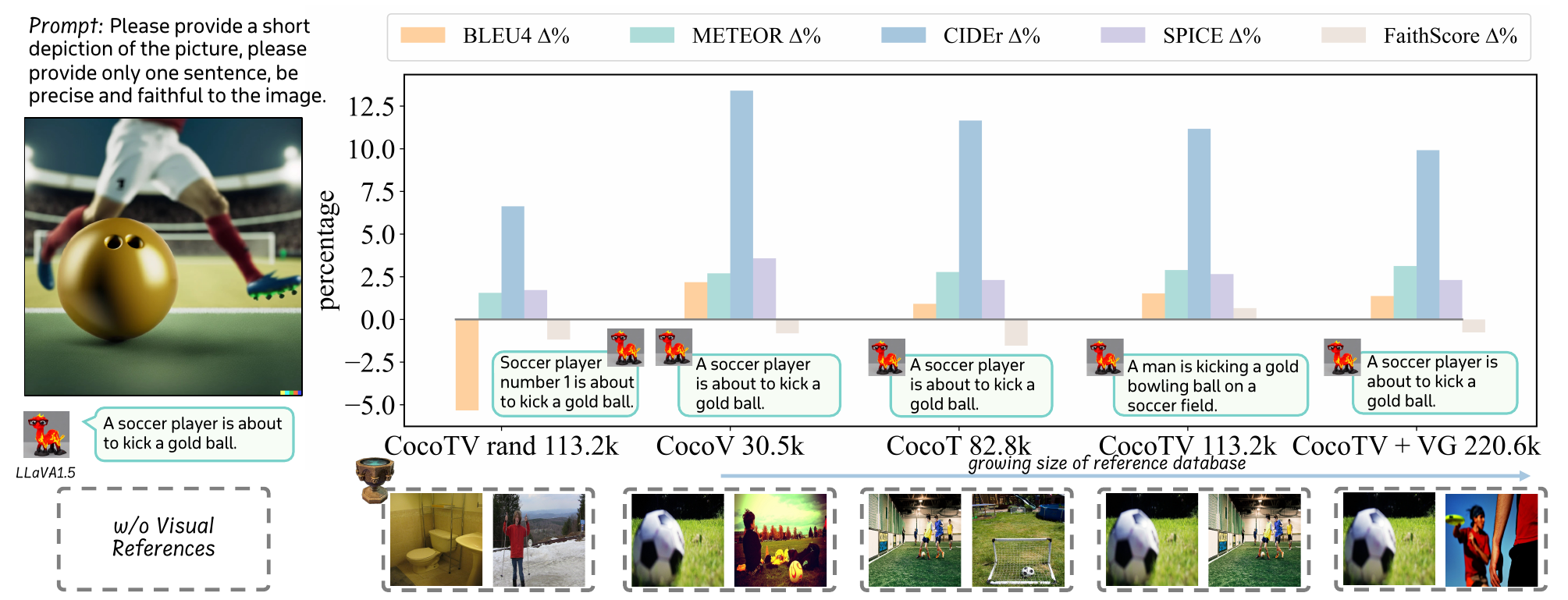}
  \caption{
    A 
    reference database containing 113k reference images from COCO Caption
    Karpathy \textit{train} (T) and \textit{restval} (V) splits
    can improve performance across all metrics.
    to address potential noisy retrieval results.
    Overall, the performance gain positively correlates with
    the similarity between the test image and the references.
    The baseline
    (depicted as the grey horizontal line)
    is Exp.1 in~\cref{tab:whoops_ablation}.
    \textit{rand} denotes sampling random references (Exp.3). 
  }
  \label{fig:ablation}
\vspace{-.35cm}
\end{figure}

We investigate the impact of the reference database's size on model performance
by incorporating more (110k) images from the Visual Genome~\cite{krishna2017visual} dataset.
Upon doubling the size,
we observe slight decrease in all metrics,
\eg, CIDEr score decreases from 75.5 to 74.6,
and more dissimilar visual content within the retrieved images
(\eg, a frisbee in~\cref{fig:ablation}).
Consequently, 
enlarging the database necessitates extra effort to filter out
noisy retrieval results.
We also validate our hypothesis that images
excluded in the training process
can enhance performance.
We build two reference databases with images from the Karpathy
\textit{train}
or \textit{restval} split only.
Notably, utilizing references merely from the \textit{restval} split 
yields the highest image captioning performance
(77.0 CIDEr score and 18.0 SPICE score,
as shown in the \textit{CocoV} column in~\cref{fig:ablation}),
outperforming the counterpart with the \textit{train} split only
(CIDEr 75.8 and SPICE 17.8).
However, combining samples from the \textit{train} and \textit{restval} splits yields the best FaithScore (68.3).

\subsubsection{Other Ablations.}
Exp.6 demonstrate that the adaptive (abbreviated adapt.) logit processing scheme
can enhance all metrics (except for the FaithScore).
Besides, increasing the number of reference images provides steady improvement in image captioning performance
(CIDEr scrore increses from 73.8 to 76.5 in Exp.7-9).
We use four references by default.

\section{Limitation}
\label{sec:limitation}

\textit{Pensieve} operates on the premise that 
in the same context,
images with similar semantics and appearance are likely to
induce analogous visual hallucinations.
We provide qualitative evidences in~\cref{fig:gen_problem}
and 
in~\cref{sect: moreExample}
to support this hypothesis.
For quantitative analysis, 
we randomly select 200 images from the OpenImages~\cite{kuznetsova2020open} validation set
and generate a one-sentence caption for each sample using LLaVA1.5-7B.
We manually review the confidence scores
within the head vocabulary
of each generated token
in all generated captions,
and find that 18 captions contain at least one hallucinatory token.
Notably, 10 out of these 18 erroneous captions
exhibit analogous visual hallucinations across similar images.
We expect 
detailed candidate-level visual hallucination annotations
and automatic evaluation pipelines
to be developed in the future.
Nonetheless, our observation in~\cref{sect:effectofnns}
demonstrates that 
analogous visual hallucinations
can be leveraged to mitigate hallucination.

\textit{Pensieve} introduces $\mathcal O (k)$ time complexity,
as distinct confidence scores are independently predicted for each visual reference.
This latency can be reduced by predicting reference confidence scores in parallel.
Note that we save
a significant amount of time and human labor 
required for data curation and model training.

\section{Conclusion}
We address 
the issue of visual hallucinations in MLLMs.
Through a comprehensive analysis, we discover that
analogous visual hallucinations induced by similar images
can be utilized to reduce hallucination.
Building on this insight,
we introduce \textit{Pensieve},
a training-free method that
allows MLLMs to retrospect similar images as references
and discern the accurate visual cues through confidence score comparison.
This 
paradigm 
corrects the
erroneous
content
that is 
mistakenly supported by MLLMs' visual or textual branches.
Quantitative and qualitative experiments on
image captioning and 
VQA
benchmarks
demonstrate the superiority of \textit{Pensieve} over other advanced decoding strategies.

\noindent\textbf{Acknowledgment:} 
This work is supported by the National Natural Science Foundation of China (No. 62372329), 
in part by the National Key Research and Development Program of China (No. 2021YFB2501104), 
in part by Shanghai Rising Star Program (No.21QC1400900), 
in part by Tongji-Qomolo Autonomous Driving Commercial Vehicle Joint Lab Project, 
and in part by Xiaomi Young Talents Program.
We would like to acknowledge the discussions and comments for this project from Zehan Zheng.

%
%
\bibliographystyle{splncs04}
\bibliography{egbib}

\clearpage
\appendix

\section{Explanations for our Analysis Pipeline}

\subsection{Semantics in MLLMs' Last Hidden State}

We term $E_c(x_i)$ the token embedding of candidate $x_i$ in the language head\footnote{
The parameters in the language head of Vicuna~\cite{zheng2023judging} is not tied to its input token embedding layer.},
where $E_c(\cdot)$ embeds token $x_i$ into a feature vector with hidden dimension $d$.
The feature vectors of all candidates 
(the total number is the length of the model's vocabulary $length(\mathcal V)$)
forms the weight matrix in the language head
with the shape of $[length(\mathcal V), d]$.
Since the language head is a linear projection layer without dropout, activate function, or bias,
the confidence score at decoding step $t$
corresponding to candidate $x_i$
is the inner product of its embedding $E_c(x_i)$ and the last hidden state $\boldsymbol{h}_t$
predicted by the MLLMs' final attention block.
If the confidence scores of different candidates are close,
then $\boldsymbol{h}_t$ is a blend of diverse semantics in near proportions.
Subsequently, those close confidence scores will result in low probabilities 
after the $\mathrm{softmax}$ operator,
indicating that MLLMs are \textbf{\textit{unsure}}
about their current prediction.

Following our proposed analysis pipeline,
if the score shift $\boldsymbol{\Delta h} \cdot E_c(x_i)$ corresponding to candidate $x_i$
is a positive value,
then $\boldsymbol{\Delta h}$ strengthens the semantics corresponding to $x_i$.
On the other hand,
some semantics 
can be weakened by $\boldsymbol{\Delta h}$,
if the candidates
get negative \textit{img score}.
If the score shift is close to zero,
we claim $\boldsymbol{\Delta h}$ is orthogonal to $E_c(x_i)$.

\subsection{Quantify MLLMs' Blindness}

We use the Jensen-Shannon Divergence (JSD) 
to help quantify 
the extent of dependency of MLLMs' predictions on the visual input.
Specifically, JSD is calculated between the confidence scores predicted by 
the test image $\boldsymbol{v}^{\tau}$ and the diffused image $\boldsymbol{v}^{d}$
in the same context $\boldsymbol{x}+\boldsymbol{y}_{<t}$,
within the head vocabulary $\mathcal V_{head}^{m}$ of $m$ top-ranked candidates.
Formally,
\begin{equation}
    \mathrm{JSD}(P \parallel Q) = \frac{1}{2} \left( D_{KL}(P \parallel M) + D_{KL}(Q \parallel M) \right)
    \label{eq:jsd1}
\end{equation}
\begin{equation}
    M = \frac{1}{2}(P + Q)
    \label{eq:jsd2}
\end{equation}
\begin{equation}
    P = \mathrm{softmax}(\{\mathit{logits}(x_j|\boldsymbol{x},\boldsymbol{y}_{<t},\boldsymbol{v}^{\tau})|x_j \in \mathcal V_{head}^{m}\})
    \label{eq:jsd3}
\end{equation}
\begin{equation}
    Q = \mathrm{softmax}(\{\mathit{logits}(x_j|\boldsymbol{x},\boldsymbol{y}_{<t},\boldsymbol{v}^{d})|x_j \in \mathcal V_{head}^{m} \})
    \label{eq:jsd4}
\end{equation}
where $D_{KL}(\cdot)$ is the KL divergence.
A high JSD value suggests that
the \textit{base scores} and the \textit{txt scores}
will yield different probability distributions following the $\mathrm{softmax}$ operator.
Therefore, the subtraction of the of \textit{txt scores} from the \textit{base scores}
(\ie, the \textit{img scores})
will \textit{\textbf{not}} be uniformly distributed,
indicating that some
(but not all) candidates
are advocated by the visual branch.
On the contrary, if both the JSD and the
first place candidate's \textit{img score} 
(denoted as the \textit{top-1 img score})
are close to zero,
then the visual information \textbf{\textit{hardly}}
impacts the prediction,
\ie, contributes few probabilities to token candidates
at the current decoding step.
From the examples in~\cref{fig:gen_problem}
and in~\cref{sect: moreExample}
(~\cref{fig:more_evidence_nocaps01} to~\cref{fig:more_evidence_whoops02}),
we notice that in a sentence,
the JSD values corresponding to erroneous
tokens in the sentence
are \textbf{\textit{not}} close to zero.
This manifests that 
MLLMs is \textbf{\textit{not}} utterly ignorant of the visual cues in the image amidst visual hallucination.

\section{Implementation Details}

\subsection{Details for the Diffusion Process.}
We implement the diffusion process~\cite{ho2020denoising} to erase visual information in the test image.
This process involves gradually adding noise to an initial image $x_{0}$
until it becomes pure Gaussian noise $x_{T}$.
At a specific step $t \in [0, T]$,
a small amount of noise is added to the image from the previous step $ t-1$. Formally,
\begin{equation}
    x_t = \sqrt{\alpha_t} x_{t-1} + \sqrt{1 - \alpha_t} \epsilon_t
    \label{eq:diffuse1}
\end{equation}
where $\alpha_t$ is a parameter that controls the amount of noise added at each step.
The parameters $\{\alpha_t\}_{t=1}^{T}$ are chosen such that they form a schedule, typically decreasing over $t$.
$\epsilon_t$ is Gaussian noise sampled from a normal distribution $\mathcal{N}(0, I)$,
where $I$ is the identity matrix.
Equation~\ref{eq:diffuse1}
is iterated for a number of steps $T$.
To obtain $x_{t}$ directly from $x_{0}$ in the forward diffusion process
without iterating through each intermediate step,
Equation~\ref{eq:diffuse2} can be derived,
which encapsulates
the cumulative effect of adding noise over multiple steps,
\begin{equation}
    x_t = \sqrt{\bar{\alpha}_t} x_0 + \sqrt{1 - \bar{\alpha}_t} \epsilon
    \label{eq:diffuse2}
\end{equation}
where $\bar{\alpha}_t = \prod_{i=1}^t \alpha_i$ is the cumulative product of the noise schedule parameters up to step $t$.
$\epsilon$ is Gaussian noise sampled from a normal distribution $\mathcal{N}(0, I)$.
We present a diffused image ($t=999$) in~\cref{fig:gen_problem},
which scarcely retains any valid visual information and is nearly indistinguishable from Gaussian noise.

\subsection{Details for Visual Retrospection}

\subsubsection{Reference Database.}
We build our reference database
with the COCO Caption~\cite{lin2014microsoft} dataset, 
which is the basis for popular visual instruction datasets~\cite{liu2023improvedllava,chen2023sharegpt4v}.
We expect the visual references to
be capable of
\textit{\textbf{inducing}} visual hallucinations,
such that the analogous hallucinations can be further leveraged to reduce hallucination.
Additionally, images in COCO Caption
contain daily and common visual concepts,
such that they 
are less likely to
induce out-of-distribution and confusing responses.
We also include the Visual Genome dataset~\cite{krishna2017visual}
to enlarge the database,
but the performance gain is lower than utilizing COCO-Caption only
(see~\cref{sec:ablation}).

\subsubsection{Retrievers.}
For image captioning, 
the retriever extracts semantic and appearance features from the visual inputs $\boldsymbol{v}$ into $E_{R}(\boldsymbol{v})$.
The CLIP~\cite{radford2021learning} vision transformer~\cite{dosovitskiy2020image} is adept at encoding semantics,
and the self-supervised pretrained DINOv2~\cite{oquab2023dinov2} ViT can capture more visual details~\cite{tong2024eyes}.
We ensemble CLIP
and
DINOv2
ViT-L14/336 model as the image retriever.
For Visual Question Answering (VQA),
The CLIP Transformer~\cite{vaswani2017attention} 
extract semantics from the question $\boldsymbol{x}$ into a vector embedding $E_{R}(\boldsymbol{x})$.
We target at finding visual references that are semantically accordant to the question.
Note that we do not add visual representations here
because they may perturb the semantics in $E_{R}(\boldsymbol{x})$.
Before text encoding, we modify the question template
in MME and
POPE
to transform questions into narratives,
\eg, replacing \textit{Is there} with \textit{A photo of}.

\subsubsection{Similarity Metrics.}

$\mathcal F (\cdot,\cdot)$ is the cosine similarity.
Specifically, representations are L2 normalized before the Maximum Inner Product Search (MIPS)
conducted by FAISS~\cite{douze2024faiss},
which is a library for vector clustering and similarity search.
For image captioning, we concatenate the $[cls]$ token from 
CLIP and DINOv2 Vision Transformer models.
Therefore, the semantic and appearance similarities are equally considered.
For VQA on the MME benchmark,
we further re-order the retrieval results based on the BLEU@1~\cite{papineni2002bleu} score
between the search query and the captions of retrieved images.


\subsection{Details for Visual Comparison}

If not otherwise specified, we refer the rank to the ranking of candidates 
according to the \textit{base scores} predicted by the test image $\boldsymbol{v}^{\tau}$.
At decoding step $t$,
we set a cut-off value as the $m^{th}$-ranked candidate's \textit{base score}.
The confidence scores lower than this threshold are set to $-inf$.
Therefore, the candidates outside the head vocabulary $\mathcal V_{head}^{m}$
will not be considered during token selection.

In the adaptive logits processing method (\cref{eq:adaptive1} to~\cref{eq:adaptive3})
the lower bound of the maximum probability after the $\mathrm{softmax}$ operator is $1/50=0.02$,
if all candidates within 
$\mathcal V_{head}^{m}$
have exactly the same confidence scores.
This scheme is designed to reduce potential erroneous support
from the diffused image $\boldsymbol{v}^{d}$
to the erroneous candidates,
especially
the visually deceptive candidates,
when $\boldsymbol{v}^{d}$ tends to equally support multiple candidates
at decoding step $t$.

\subsubsection{Compare when Necessary.}
\label{sect:jsd_thres}
We observe that the JSD value can be very close to zero at some decoding steps,
especially in subwords
and lexical collocation,
such as $ing$ after $\_lay$ in~\cref{fig:more_evidence_nocaps01},
and $ck$ after $\_du$ in~\cref{fig:more_evidence_whoops02}.
Modifying their confidence score distribution
may harm language fluency and induce grammatical errors.
Therefore, we optionally set a JSD threshold to disable \textit{Pensieve} 
at certain decoding steps
(although we do not use it in practice for image captioning in all reported quantitative experiments).

\section{Experiments}

\subsection{Detailed Experimental Settings}

\subsubsection{Datasets and Metrics.}

We use POPE
(the Polling-based Object Probing Evaluation~\cite{li2023evaluating}),
the Count and Existence subtasks of MME~\cite{fu2023mme}
for object-level visual hallucination evaluation.
The Color and Position subtasks of MME
assess the attribute-level hallucination.
On these benchmarks, 
The MLLMs are explicitly prompted to answer yes or no.
Based on whether the response contains the string $yes$ or $no$, 
the Accuracy, Precision, Recall, and F1 score are determined on POPE,
and the Accuracy and Accuracy$+$ are combined on MME
as the official evaluation metrics, respectively.

Considering that the performance 
in binary VQA task may not
be able to
capture the extent of
visual hallucination in free-form text generation process,
we further evaluate our proposed \textit{Pensieve}
on two challenging benchmarks.
The Whoops~\cite{bitton2023breaking} benchmark
contains 500 generated high-fidelity images with 
some of the visual cues violating common sense.
LLaVA Bench in the Wild~\cite{liu2024visual}(abbreviated as $LLaVA^W$)
contains 60 questions on 24 images,
including indoor and outdoor scenes, memes, paintings, and sketches.

For visual hallucination evaluation on Whoops
and $LLaVA^{W}$,
we use the FaithScore~\cite{jing2023faithscore} metric,
which is adept at
judging the extent of visual hallucinations
at a finer granularity
(beyond object and attribute-level errors),
and shows a higher correlation with human judgment
than traditional metrics
like CHAIR~\cite{rohrbach2018object} 
and CLIP-Score~\cite{hessel2021clipscore}.
Specifically, 
We prompt ChatGPT
(the \textit{gpt-3.5-turbo-1106} API\footnote{\url{https://openai.com/blog/chatgpt}})
to identify descriptive sentences in the response,
and extract atomic facts from those sentences.
For atomic facts verification,
we use the recommended model
OFA~\cite{wang2022ofa},
instead of a model from the LLaVA or BLIP family,
as we are evaluating LLaVA-1.5-7B
and InstructBLIP-7B.
OFA is finetuned on the visual entailment dataset SNLI-VE~\cite{xie2019visual},
showing comparable performance as LLaVA-1.5 on atomic fact verification~\cite{jing2023faithscore}.
The FaithScore 
is adept at indicating the comprehensive degree of visual hallucination.
For all experiments, 
we query ChatGPT three times and report the averaged results.

\subsubsection{Hyper-parameters}
\label{sect:hyper_param_setting}
are detailed in~\cref{tab:hyperparam}.
Note that for InstructBLIP on the POPE-GQA subset only,
the hyper-parameters are set to
$\alpha_{\tau}=1.0$, $\beta_{d}=0.1$, $\beta_{NN}=0.1$
for better performance.

\begin{table}[tb]
  \centering
  \scalebox{0.8}{
  \begin{tabular}{cccccc}
    \toprule
    Model & Hyper-param. & Whoops & LLaVA$^{W}$ & MME & POPE  \\
    \midrule
    \multirow{5}*{LLaVA-1.5} & $k$ & 4 & 2 & 1 & 2 \\
    ~ & $\beta_{NN}$ & 0.1 & 0.1 & 0.01 & 0.05 \\
    ~ & $\beta_{d}$ & 0.1 & 0.1 & 0.1 & 0.01 \\
    ~ & $\alpha_{\tau}$ & 1.0 & 1.0 & 1.0 & 1.0 \\
    ~ & diffu. step & 900 & 900 & 700 & 900 \\
    \cline{2-6}
    \multirow{5}*{InstructBLIP} & $k$ & 2 & 2 & 2 & 2 \\
    ~ & $\beta_{NN}$ & 0.04 & 0.1 & 0.1 & 0.02 \\
    ~ & $\beta_{d}$ & 0.05 & 0.1 & 0.5 & 0.02 \\
    ~ & $\alpha_{\tau}$ & 1.2 & 1.0 & 1.0 & 1.5 \\
    ~ & diffu. step & 500 & 900 & 900 & 900 \\
  \bottomrule
  \end{tabular}}
  \caption{
    Detailed hyper-parameter setting for LLaVA-1.5
    and InstructBLIP
    for each test dataset.
    diffu. step denotes the image diffusion step.
  }
  \label{tab:hyperparam}
\end{table}

\subsubsection{MLLM Baselines.}

We select two representative open-source MLLMs~\cite{liu2023improvedllava,instructblip} to implement our method.
Both MLLMs integrate CLIP Vision Transformer as the visual encoder.
As for the cross-modal connector, 
LLaVA-1.5 uses a two-layer MLP,
and InstructBLIP uses a Q-former~\cite{li2023blip} with textual input.
We use the 7B version of both MLLMs,
with Vicuna-7B~\cite{zheng2023judging} as the language decoder.

\subsubsection{Settings for the Ablation Study.}

For all ablation studies,
we use LLaVA-1.5-7B as the baseline MLLM,
and the default hyper-parameter setting 
(the same as the experiment on Whoops).
In ablation study Exp.3, we retrieve five sets of random visual references for each test sample,
and report the averaged results on five separate runs.
In Exp.4, we set $\beta_{d}=1e-6$ to discard the impact of the diffused image $\boldsymbol{v}^{d}$
on confidence score modification.
We do not discard the diffused image itself,
since we still need it to calculate $\alpha_{d}^{t}$ and $\alpha_{NN}^{t}$. 
In Exp.6, we fix both $\alpha_{d}^{t}$ and $\alpha_{NN}^{t}$ to 0.1.
In the study of the reference database size,
we include all images (108k) from the Visual Genome~\cite{krishna2017visual} dataset.

\subsection{More Results on LLaVA Bench in the Wild}


\begin{figure}[h]
  \centering
  \includegraphics[width=11cm]{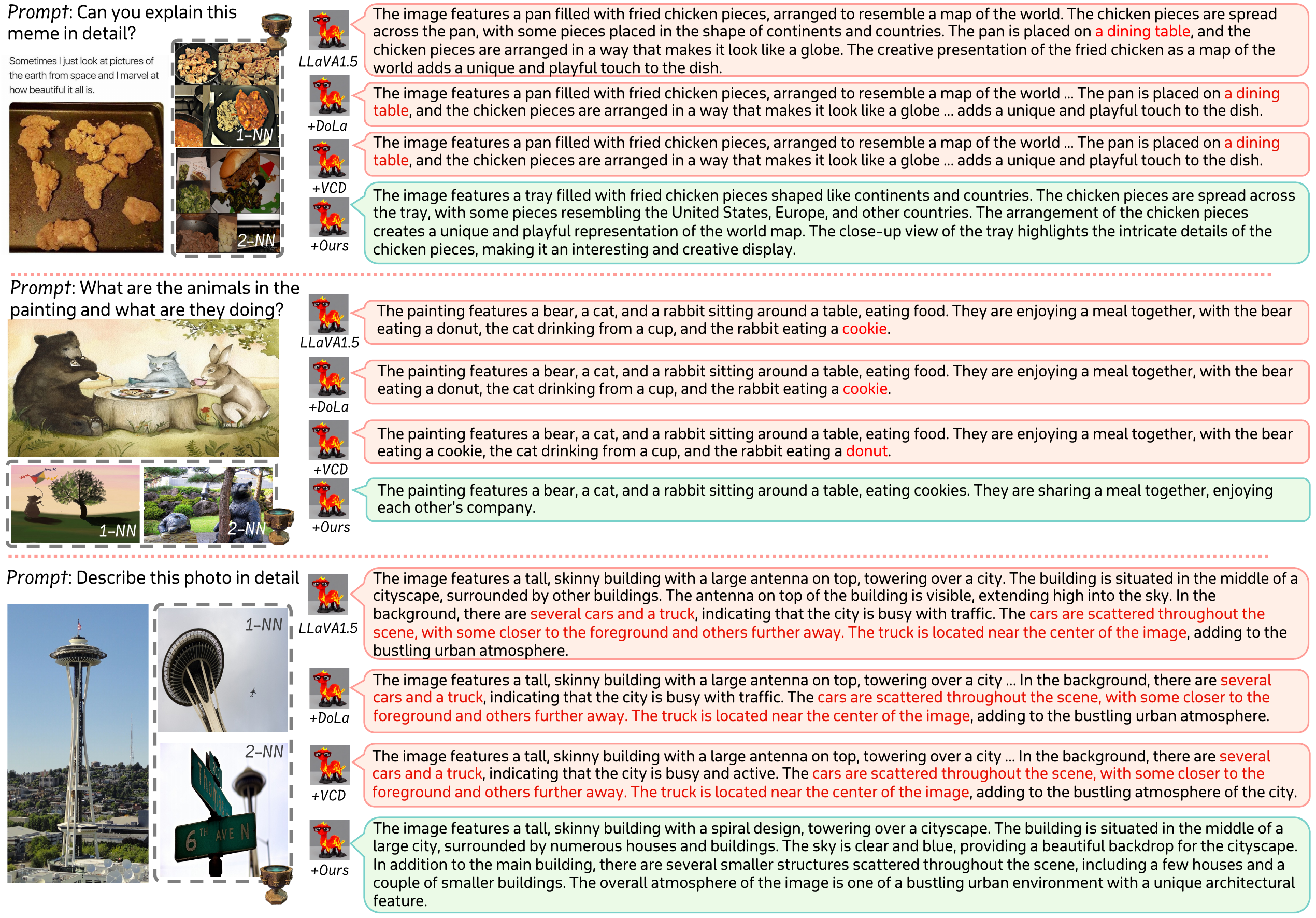}
  \caption{
    More qualitative results on LLaVA-Bench in the wild.
    Our proposed \textit{Pensieve} effectively reduces visual hallucination for
    LLaVA1.5-7B.
    Retrieved visual references are
    in the grey box.
    Hallucinatory content is marked in red.
  }
  \label{fig:llavabench_new_llava}
\end{figure}

\begin{figure}[h]
  \centering
  \includegraphics[width=11cm]{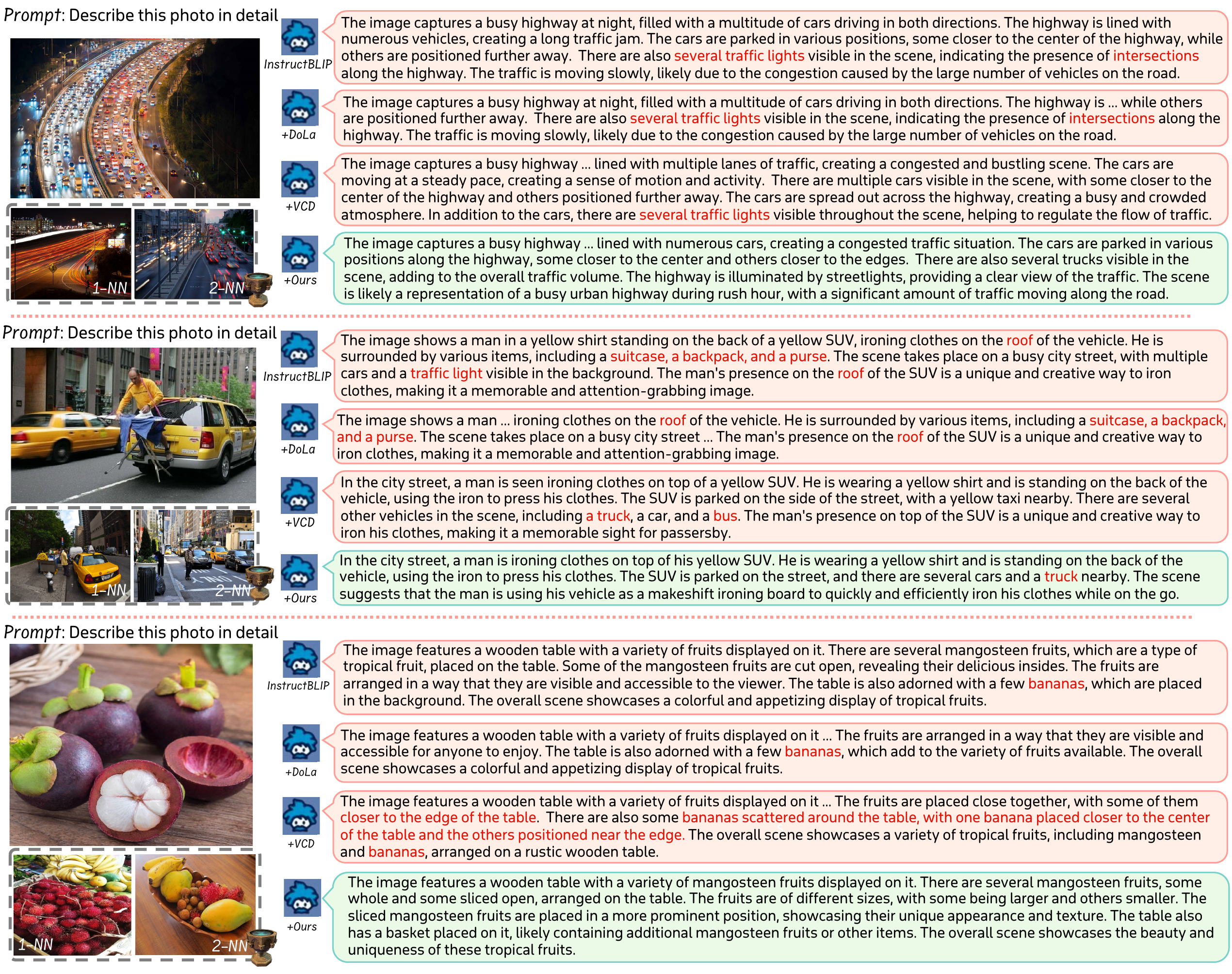}
  \caption{
    More qualitative results on LLaVA-Bench in the wild.
    Our proposed \textit{Pensieve} effectively reduces visual hallucination for
    InstructBLIP-7B.
    Retrieved visual references are
    in the grey box.
    Hallucinatory content is marked in red.
  }
  \label{fig:llavabench_new_ib}
\end{figure}

We present more qualitative results
to demonstrate the superiority of
our proposed \textit{Pensieve}.
Figures~\ref{fig:llavabench_new_llava} and~\ref{fig:llavabench_new_ib}
demonstrate that
\textit{Pensieve} helps LLaVA1.5 and InstructBLIP 
correctly describe challenging scenes,
which contain visual cues that are difficult to distinguish.
Specifically,
\textit{Pensieve} helps
LLaVA1.5 correctly identify the small buildings scattered around the large antenna,
and helps InstructBLIP discern the streetlights from traffic lights
near the highway.
\textit{Pensieve} also successfully corrects visual hallucinations that 
confuse both DoLa
and VCD
such as the dining table,
the bananas, 
and the man ironing on the roof of a car.
Additionally,
\textit{Pensieve} is capable of
correcting erroneous 
nouns, adjectives, prepositions, \textit{etc.,}
wherever
they appear in the sentence.

In summary,
\textit{Pensieve}
generalizes well on various image domains,
including photographs, paintings, and text-rich images.
Moreover, 
\textit{Pensieve} is a versatile method that
is adept at mitigating various
hallucination categories,
\textbf{NOT} limited to pre-defined error types.

\subsection{More Results on MME}

\subsubsection{Quantitative Results.}

We present results on all perception tasks of MME in~\cref{tab:mme_supple}.
Note that
we concentrate on the problem of visual hallucination,
facilitating MLLMs to faithfully describe the image.
The cognition tasks that necessitate specified knowledge
(commonsense reasoning, numerical calculation, text translation, and code reasoning)
is beyond the scope of this work.
Table~\ref{tab:mme_supple} demonstrate that our proposed \textit{Pensieve}
improves the overall perception performance for both MLLMs,
increasing the total score for LLaVA1.5 by 56.6,
and the total score for InstructBLIP by 76.3,
outperforming other advanced decoding strategies.

\begin{table}[h]
\renewcommand\arraystretch{1.0}
  \caption{
    Results on all perception sub-tasks of the MME benchmark.
    We report the officially defined metric that combines Accuracy and Accuracy+.
    \textit{Pensieve} substantially improves the perception competencies for
    both MLLMs.
  }
  \label{tab:mme_supple}
  \centering
  \scalebox{0.6}{
  \begin{tabular}{p{2.0cm}p{1.6cm}p{1.2cm}p{1.3cm}p{1.6cm}p{1.4cm}p{1.3cm}p{1.6cm}p{1.2cm}p{1.7cm}p{1.5cm}p{1.2cm}p{1.3cm}}
    \toprule
    Model & Decoding & Color $\uparrow$ & Count $\uparrow$ & Existence $\uparrow$ & Position $\uparrow$ & Posters $\uparrow$ & Celebrity $\uparrow$ & Scene $\uparrow$ & Landmark $\uparrow$ & Artwork $\uparrow$ & OCR $\uparrow$ & Total $\uparrow$\\
    \midrule
    \multirow{4}*{LLaVA-1.5} & greedy & 155.0 & \bf 158.3 & 195.0 & 123.3 & 129.6 & 132.6 & 155.0 & 163.5 & 121.0 & 125.0 & 1458.9 \\
    ~ & +{\it DoLa} & 153.3 & 158.3 & 195.0 & 123.3 & 127.6 & 130.9 & 154.8 & 162.8 & 122.3 & 122.5 & 1450.7 \\
    ~ & +{\it VCD} & 148.3 & 158.3 & 190.0 & 126.7 & 136.7 & 147.4 & 148.8 & \bf 166.0 & 122.5 & 130.0 & 1474.7 \\
    ~ & +{\it Ours} & \bf 165.0 & 153.3 & \bf 195.0 & \bf 128.3 & \bf 141.8 & \bf 150.3 & \bf 157.3 & 161.8 & \bf 122.8 & \bf 140.0 & \bf 1515.5 \\
    \cline{2-13}
    \multirow{4}*{InstructBLIP} & greedy & 120.0 & 60.0 & \bf 185.0 & 50.0 & 142.9 & 81.8 & 160.0 & 160.0 & 92.0 & 65.0 & 1116.6 \\
    ~ & +{\it DoLa} & 120.0 & 60.0 & 185.0 & 50.0 & 142.9 & 80.9 & 160.0 & 160.0 & 92.2 & 65.0 & 1116.0 \\
    ~ & +{\it VCD} & 123.3 & 60.0 & 185.0 & 53.3 & \bf 151.7 & \bf 94.1 & 156.5 & \bf 161.3 & \bf 99.3 & 95.0 & 1179.5 \\
    ~ & +{\it Ours} & \bf 153.3 & \bf 78.3 & 180.0 & \bf 58.3 & 140.5 & 71.2 & \bf 163.8 & 158.3 & 94.3 & \bf 95.0 & \bf 1192.9 \\
  \bottomrule
  \end{tabular}}
\end{table}

\subsubsection{Case Study.}
\label{sec:case_study_mme}
\begin{figure}[!p]
  \centering
  \includegraphics[width=11cm]{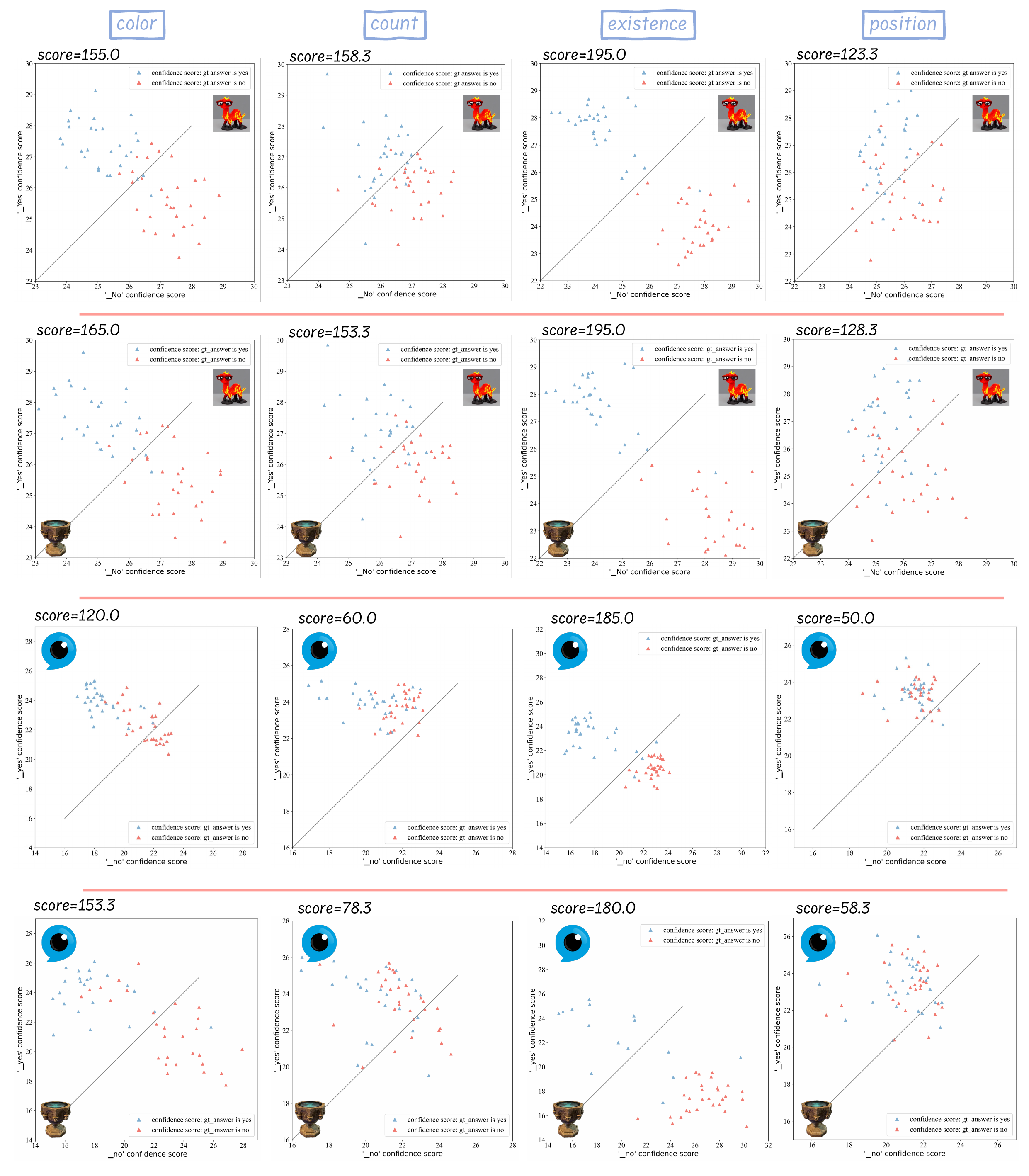}
  \caption{
    Case study for the impact of \textit{Pensieve} on the MME benchmark.
    We plot the confidence scores
    corresponding to the candidates $\_yes$ and $\_no$
    on the vertical and horizontal axis, respectively.
    Each test sample corresponds to
    a blue or red triangle mark,
    indicating that the ground truth answer
    is $\_yes$ or $\_no$, respectively.
    Our proposed \textit{Pensieve} reduces false positives
    on four hallucination subtasks of the MME benchmark.
    Best viewed in color and zoomed in.
  }
  \label{fig:mme_qualitative}
\end{figure}

We investigate how \textit{Pensieve} enhances model performance
on the MME benchmark
and present the results in~\cref{fig:mme_qualitative}.
We plot the confidence scores
corresponding to the candidates $\_yes$ and $\_no$
on the vertical and horizontal axis, respectively.
Each test sample corresponds to
a single triangle mark
colored in blue
(the samples' ground truth answer is $\_yes$)
or red
(the ground truth answer is $\_no$).
If a blue triangle mark is below the $y=x$ line, 
or a red triangle mark is above the $y=x$ line, 
then the answer is incorrect.
We summarize our main observations as follows:

\begin{itemize}
    \item Without visual references,
    the scores of the two candidates can be close.
    As illustrated in the first row in~\cref{fig:mme_qualitative},
    a number of predicted samples in the Count and Position task
    are close to the $y=x$ line.
    This indicates LLaVA1.5 has equal confidence,
    \ie, high perplexity,
    to the positive and negative answers.
    
    \item InstructBLIP tends to answer yes
    for all provided questions,
    regardless of their ground truth label.
    As shown in the third row in~\cref{fig:mme_qualitative},
    almost all samples in the Count and Position task are located above the $y=x$ line.
\end{itemize}

In our paradigm,
MLLMs are allowed to retrospect visual references that are semantically accordant to the image.
We expect the comparison with such \textit{standard answers} can help reduce false positives.
Figure~\ref{fig:mme_qualitative} illustrates that after integrating \textit{Pensieve}
(in the second and fourth row),
there are fewer red triangle marks
above the $y=x$ line,
\ie, less false positives,
especially in the Color subtask.
This indicates that the number of false positives is reduced.

\subsection{More Results on POPE}

Detailed results on three official splits of POPE benchmark 
(the random, popular, and adversarial splits)
are shown in~\cref{tab:pope_supple}.
\textit{Pensieve} boosts the accuracy and the F1 score for
both LLaVA1.5
and InstructBLIP
on almost all splits of all subsets
(COCO, AOKVQA and GQA),
exceeding DoLa and VCD in
varying degrees.
We notice
\textit{Pensieve}'s impact on 
InstructBLIP slightly oscillates in the COCO subset,
while the performance is consistently better than
VCD.

\begin{table}[h]
  \caption{
    Results on the POPE benchmark MSCOCO, AOKVQA and GQA subsets.
    We report the performance in random, popular, and adversarial settings.
    \textit{Pensieve} improves the overall performance for both MLLMs.
  }
  \label{tab:pope_supple}
  \centering
  \scalebox{0.78}{
  \begin{tabular}{p{1.8cm}p{2.2cm}p{1.6cm}p{1.4cm}p{1.4cm}p{1.4cm}p{1.4cm}p{1.4cm}p{1.4cm}}
    \toprule
    \multirow{2}*{Setting} & \multirow{2}*{Model} & \multirow{2}*{Decoding} & \multicolumn{2}{c}{\cellcolor{gray!20}COCO} & \multicolumn{2}{c}{\cellcolor{gray!40}AOKVQA} & \multicolumn{2}{c}{\cellcolor{gray!20}GQA}\\
    ~ & ~ & ~ & {\bf Acc. $\uparrow$} & {\bf F1 $\uparrow$} & {\bf Acc. $\uparrow$} & {\bf F1 $\uparrow$} & {\bf Acc. $\uparrow$} & {\bf F1 $\uparrow$} \\
    \midrule
    \multirow{8}*{Random} & \multirow{4}*{LLaVA-1.5} & greedy & 87.13 & 85.58 & 88.73 & 88.16 & 89.33 & 88.81 \\
    ~ & ~ & +{\it DoLa} & 86.93 & 85.31 & 88.66 & 88.06 & 89.30 & 88.74 \\
    ~ & ~ & +{\it VCD} & \bf 88.63 & \bf 88.15 & 87.16 & 87.46 & 88.03 & 88.43 \\
    ~ & ~ & +{\it Ours} & 87.53 & 86.17 & \bf 89.20 & \bf 88.84 & \bf 89.46 & \bf 89.16 \\
    \cline{2-9}
    ~ & \multirow{4}*{InstructBLIP} & greedy & 87.97 & 86.97 & 88.50 & 88.54 & \bf 87.26 & \bf 87.31 \\
    ~ & ~ & +{\it DoLa} & \bf 88.00 & \bf 87.03 & 88.50 & \bf 88.57 & 87.23 & 87.29 \\
    ~ & ~ & +{\it VCD} & 87.07 & 85.97 & 86.80 & 86.97 & 86.37 & 86.38 \\
    ~ & ~ & +{\it Ours} & 87.83 & 86.79 & \bf 88.53 & 88.54 & 87.23 & 86.70 \\
    \midrule
    \multirow{8}*{Popular} & \multirow{4}*{LLaVA-1.5} & greedy & 85.90 & 84.41 & 85.30 & 85.09 & 84.00 & 84.12 \\
    ~ & ~ & +{\it DoLa} & 85.70 & 84.14 & 85.30 & 85.05 & 84.00 & 84.05 \\
    ~ & ~ & +{\it VCD} & 86.13 & \bf 85.93 & 83.07 & 83.94 & 82.43 & 83.85 \\
    ~ & ~ & +{\it Ours} & \bf 86.13 & 84.87 & \bf 85.57 & \bf 85.62 & \bf 84.53 & \bf 84.89 \\
    \cline{2-9}
    ~ & \multirow{4}*{InstructBLIP} & greedy & 84.97 & 84.24 & 81.86 & 83.05 & 78.60 & 80.37 \\
    ~ & ~ & +{\it DoLa} & \bf 85.06 & \bf 84.36 & 81.86 & 83.09 & 78.57 & 80.37 \\
    ~ & ~ & +{\it VCD} & 84.43 & 83.62 & 81.63 & 82.77 & 79.87 & \bf 81.15 \\
    ~ & ~ & +{\it Ours} & 84.90 & 84.12 & \bf 82.03 & \bf 83.16 & \bf 80.20 & 80.78 \\
    \midrule
    \multirow{8}*{Adversarial} & \multirow{4}*{LLaVA-1.5} & greedy & 83.63 & 82.33 & 78.93 & 79.94 & 80.87 & 81.58 \\
    ~ & ~ & +{\it DoLa} & 83.50 & 82.12 & \bf 78.93 & 79.87 & 80.83 & 81.48 \\
    ~ & ~ & +{\it VCD} & 82.00 & 82.51 & 75.96 & 78.75 & 78.03 & 80.62 \\
    ~ & ~ & +{\it Ours} & \bf 83.73 & \bf 82.70  & 78.90 & \bf 80.29 & \bf 81.00 & \bf 82.09 \\
    \cline{2-9}
    ~ & \multirow{4}*{InstructBLIP} & greedy & 82.50 & 82.15  & 74.83 & 77.93 & 75.87 & 78.40 \\
    ~ & ~ & +{\it DoLa} & 82.56 & \bf 82.24 & 74.86 & 78.00 & 75.86 & 78.42 \\
    ~ & ~ & +{\it VCD} & 81.77 & 81.27 & \bf 76.06 & \bf 78.61 & 76.47 & 78.63 \\
    ~ & ~ & +{\it Ours} & \bf 82.63 & 82.21 & 74.93 & 77.96 & \bf 77.83 & \bf 78.98 \\
  \bottomrule
  \end{tabular}}
\end{table}

\subsection{More Ablations}

\subsubsection{Token Sampling.}

In the main paper,
greedy search is used as the baseline decoding strategy,
\ie, always selecting the candidate with the highest confidence score.
We further study the integration of \textit{Pensieve}
with different token sampling strategies in~\cref{tab:token_sample_whoops}.
Greedy search avoids errors induced by token sampling and thus has the highest performance.
Our proposed \textit{Pensieve} can be well integrated with various sampling strategies,
enhancing image captioning performance and reducing visual hallucination.
For nucleus (Top-p) sampling, we set $p = 0.9$. 
For Top-k sampling, we set $k = 50$.
We fix the random seeds in all experiments.

\begin{table}[h]
\renewcommand\arraystretch{1.02}
  \caption{
    Token sampling experiments on Whoops
    benchmark.
    \textit{Pensieve} enhances the performance for 
    LLaVA-1.5
    across various baseline decoding strategies, including
    greedy search and different token sampling methods.
  }
  \label{tab:token_sample_whoops}
  \centering
  \scalebox{0.8}{
  \begin{tabular}{p{3.7cm}p{1.4cm}p{1.1cm}p{1.1cm}p{1.1cm}p{1.1cm}p{1.1cm}}
    \toprule
    \multicolumn{2}{c}{Method} & B4 $\uparrow$ & M $\uparrow$ & C $\uparrow$ & S $\uparrow$ & FS\% $\uparrow$\\
    \midrule
    \rowcolor{gray!20}
    \multicolumn{7}{c}{\textbf{\textit{zeroshot}}} \\
    \multirow{8}*{LLaVA-1.5 \scalebox{0.8}{\color{gray}{Vicuna-7B}}} & greedy & 19.7 & 25.6 & 67.9 & 17.3 & 67.9 \\
    ~ & +{\it Ours} & \bf 20.0 & \bf 26.3 & \bf 75.5 & \bf 17.8 & \bf 68.3 \\
    \cline{2-7}
    ~ & sample &  6.9 & 19.0 & 31.4 & 12.0 & 57.8 \\
    ~ & +{\it Ours} & 8.8 & 20.3 & 40.6 & 13.7 & 61.9 \\
    \cline{2-7}
    ~ & nucleus & 9.6 & 20.5 & 39.4 & 13.3 & 62.4 \\
    ~ & +{\it Ours} & 10.7 & 21.8 & 46.8 & 14.0 & 65.4 \\
    \cline{2-7}
    ~ & Top-k & 8.0 & 19.3 & 33.6 & 12.3 & 59.1 \\
    ~ & +{\it Ours} & 8.8 & 20.3 & 40.6 & 13.7 & 62.4 \\
  \bottomrule
  \end{tabular}}
\end{table}

\section{More Examples to Support our Premise}
\label{sect: moreExample}

In this section,
we provide more examples
to support our claim that MLLMs may not be utterly ignorant of accurate visual cues when they hallucinate.
Moreover,
images with similar semantics and appearance can
induce analogous visual hallucinations in the same context,
and token candidates with sharper variation in confidence score distribution
are more likely to be the accurate ones.

\begin{figure}[!p]
  \centering
  \includegraphics[width=11cm]{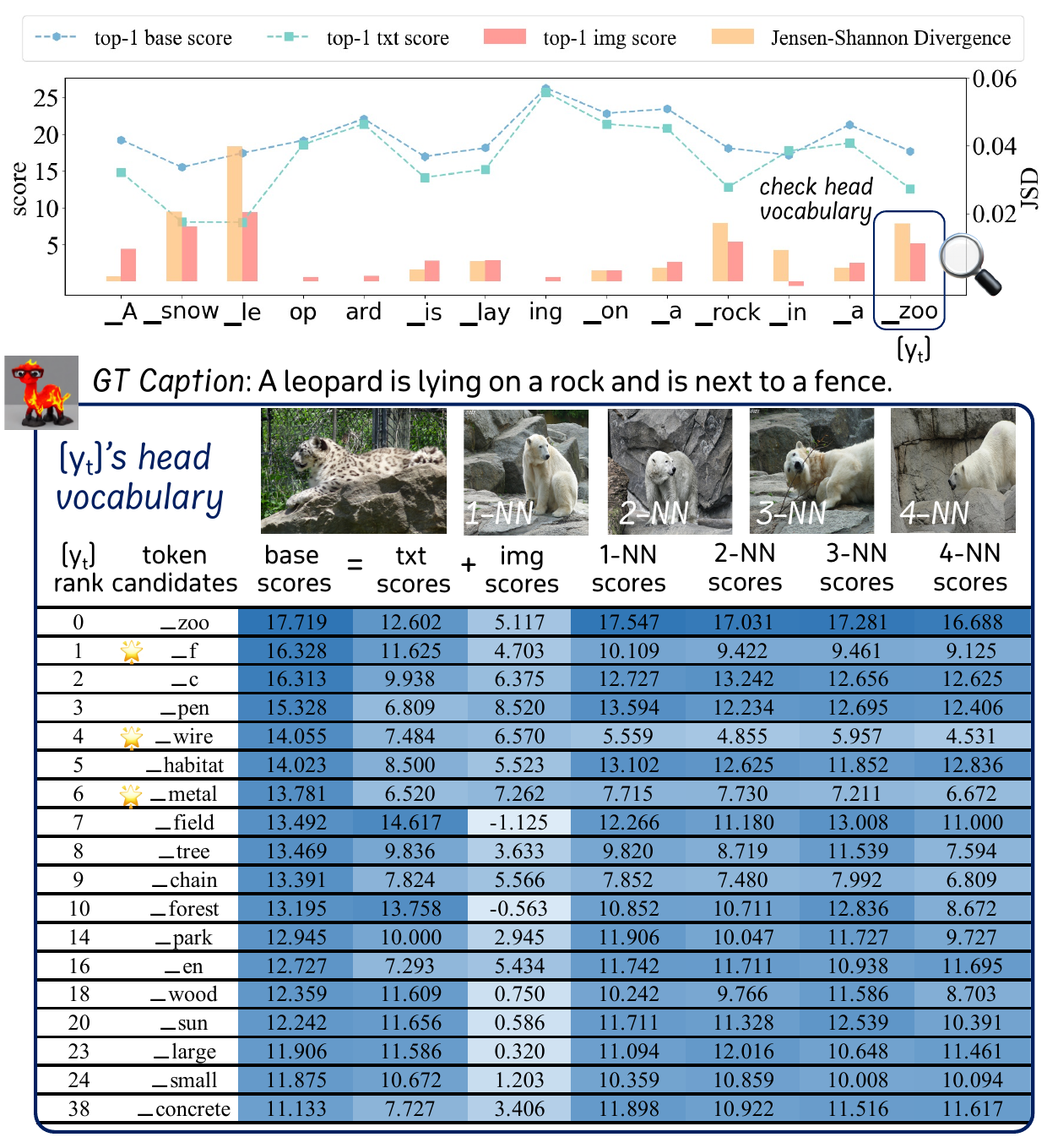}
  \caption{
    More qualitative evidence to support our hypothesis.
    This caption is predicted by LLaVA1.5-7B.
    The test image is from the OpenImages validation set,
    and the references are from the COCO Caption dataset.
    The star emojis indicate the accurate candidates.
  }
  \label{fig:more_evidence_nocaps01}
\end{figure}

\begin{figure}[!p]
  \centering
  \includegraphics[width=11cm]{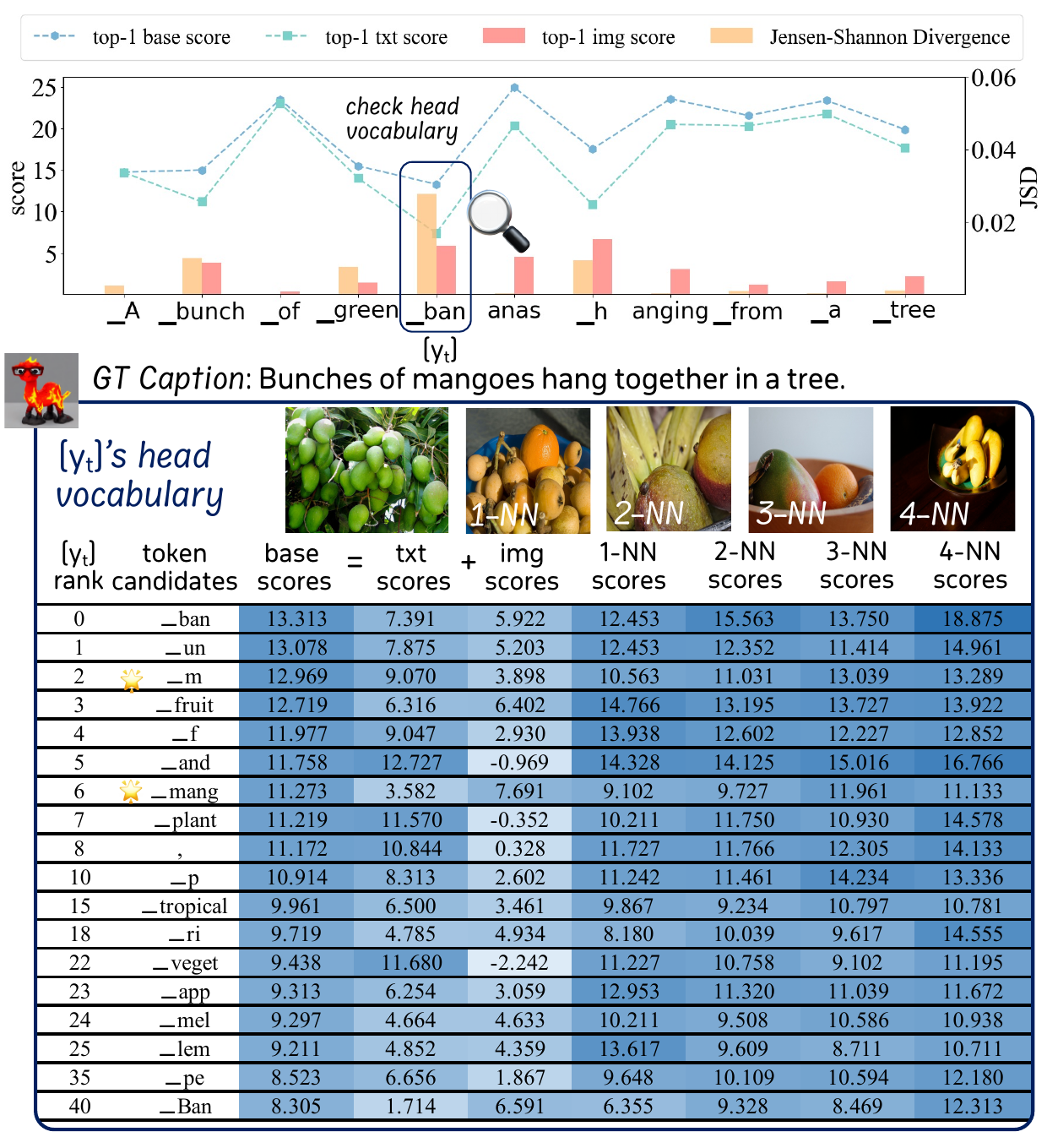}
  \caption{
    More qualitative evidence to support our hypothesis.
    This caption is predicted by LLaVA1.5-7B.
    The test image is from the OpenImages validation set,
    and the references are from the COCO Caption dataset.
    The star emojis indicate the accurate candidates.
  }
  \label{fig:more_evidence_nocaps02}
\end{figure}

We first provide two examples with the test image from the OpenImages~\cite{kuznetsova2020open} validation set in~\cref{fig:more_evidence_nocaps01} and~\cref{fig:more_evidence_nocaps02}.
For the hallucinated content
$\_zoo$ and $\_ban$,
which ranks first
in the vocabulary,
the visual branch contributed +5.117 and +5.922 \textit{img scores}, respectively.
Meanwhile, the visual modality also
contributes comparable \textit{img scores} for the accurate candidates,
\eg, +4.703 scores for $\_f$ (the first token for fence, ranks second),
and +3.898 for $\_m$ (the first token for mango, ranks third).
This indicates that the MLLM is aware of the accurate visual cues amidst hallucination,
whereas the visual branch also mistakenly advocated other erroneous candidates.
In these scenarios, 
directly
implementing VCD
may further exacerbate visual hallucinations,
as the \textit{img scores} of the hallucinatory candidates are higher than that of the accurate ones.

In~\cref{fig:more_evidence_nocaps01} and~\cref{fig:more_evidence_nocaps02},
we also observe analogous visual hallucinations among similar images,
\eg, $\_zoo$, $\_c$ (first token for cage), and $\_pen$
in Figure~\ref{fig:more_evidence_nocaps01},
as well as
$\_ban$ and $\_un$ (first token for unripe) in Figure~\cref{fig:more_evidence_nocaps02}.
These hallucinatory candidates obtain relatively high confidence scores 
even though the input images \textit{\textbf{do not}} contain such
visual information.
Moreover,
another notable phenomenon emerges:
the accurate candidates'
\textit{kNN scores}
are significantly lower
than that of the erroneous candidates,
\eg, the accurate candidate $\_f$'s
confidence score dropped from 16.328 to 9.529 in average.
This change is more significant than the hallucinatory candidates,
\eg, $\_zoo$ from 17.719 to 17.137,
$\_c$ from 16.313 to 12.813,
and $\_pen$ from 15.328 to 12.732.
This difference can help distinguish the accurate content from hallucinations.

\begin{figure}[!p]
  \centering
  \includegraphics[width=11cm]{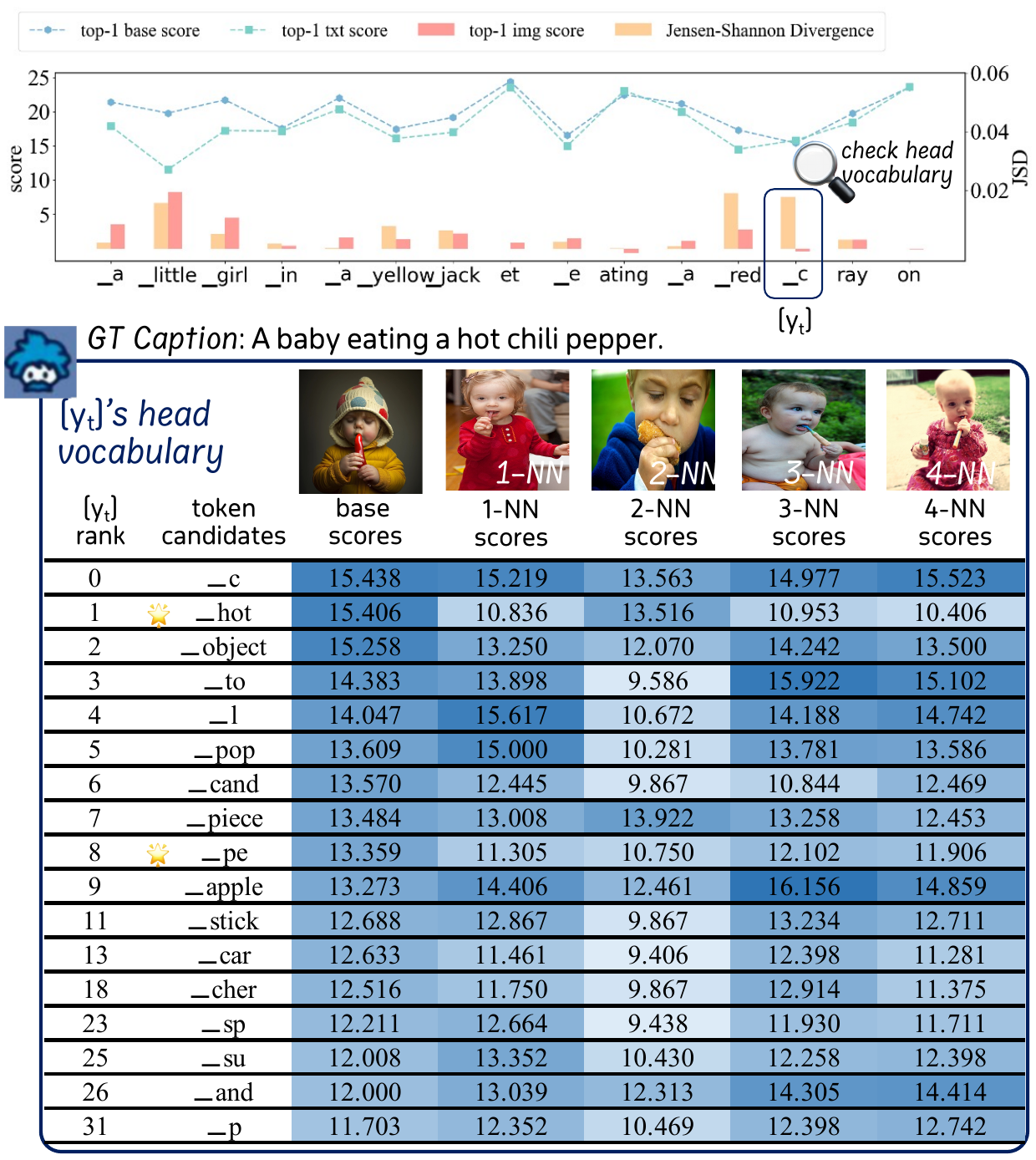}
  \caption{
    More qualitative evidence to support our hypothesis.
    This caption is predicted by InstructBLIP-Vicuna-7B.
    The test image is from the Whoops benchmark,
    and the references are from the COCO Caption dataset.
    The star emojis indicate the accurate candidates.
  }
  \label{fig:more_evidence_whoops01}
\end{figure}

\begin{figure}[!p]
  \centering
  \includegraphics[width=11cm]{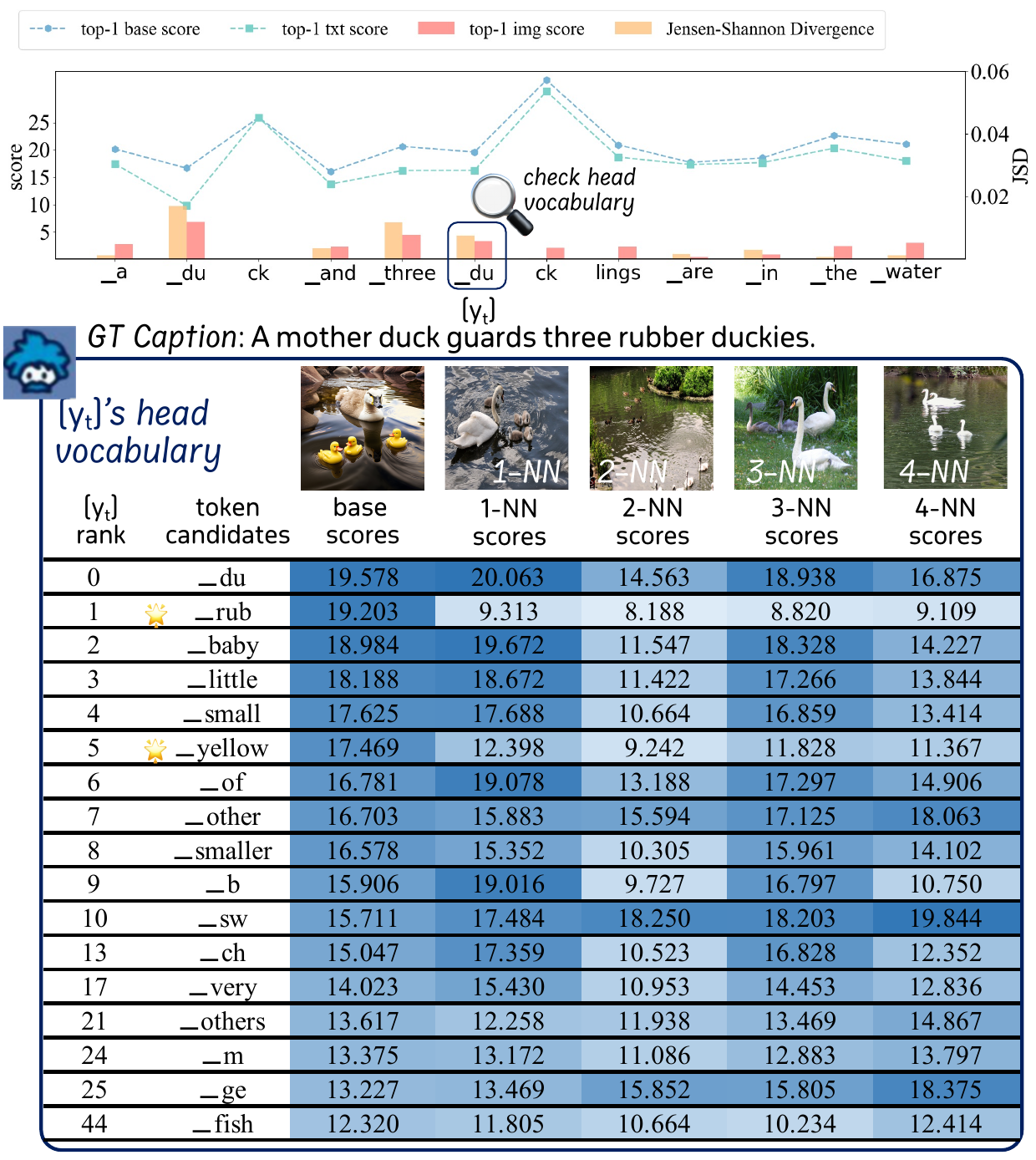}
  \caption{
    More qualitative evidence to support our hypothesis.
    This caption is predicted by InstructBLIP-Vicuna-7B.
    The test image is from the Whoops benchmark,
    and the references are from the COCO Caption dataset.
    The star emojis indicate the accurate candidates.
  }
  \label{fig:more_evidence_whoops02}
\end{figure}

\begin{figure}[!p]
  \centering
  \includegraphics[width=11cm]{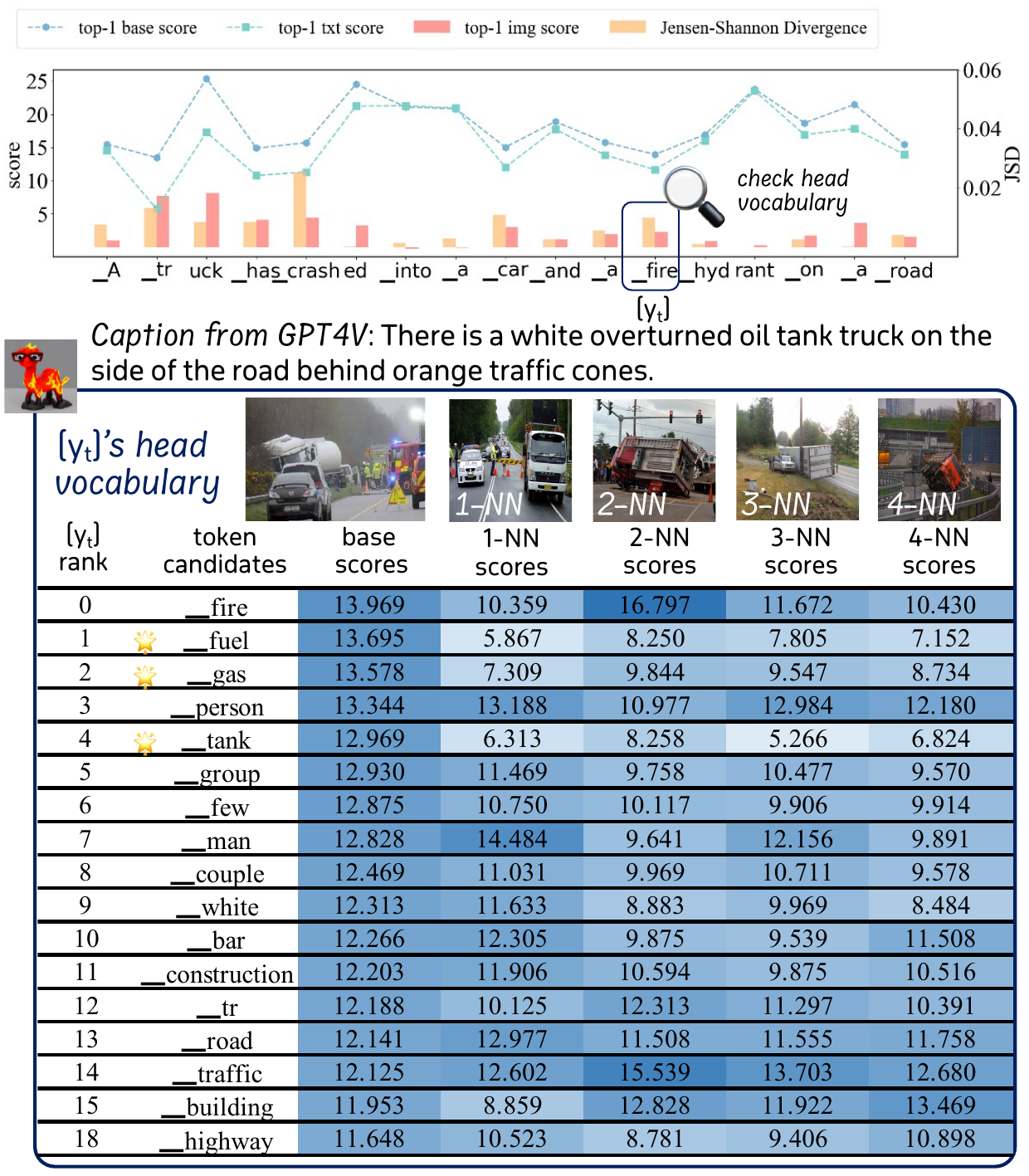}
  \caption{
    More qualitative evidence to support our hypothesis.
    This caption is predicted by LLaVA1.5-7B.
    The test image is crawled from the internet,
    and the references are from the COCO Caption dataset.
    The star emojis indicate the accurate candidates.
  }
  \label{fig:more_evidence_web01}
\end{figure}

\begin{figure}[!p]
  \centering
  \includegraphics[width=11cm]{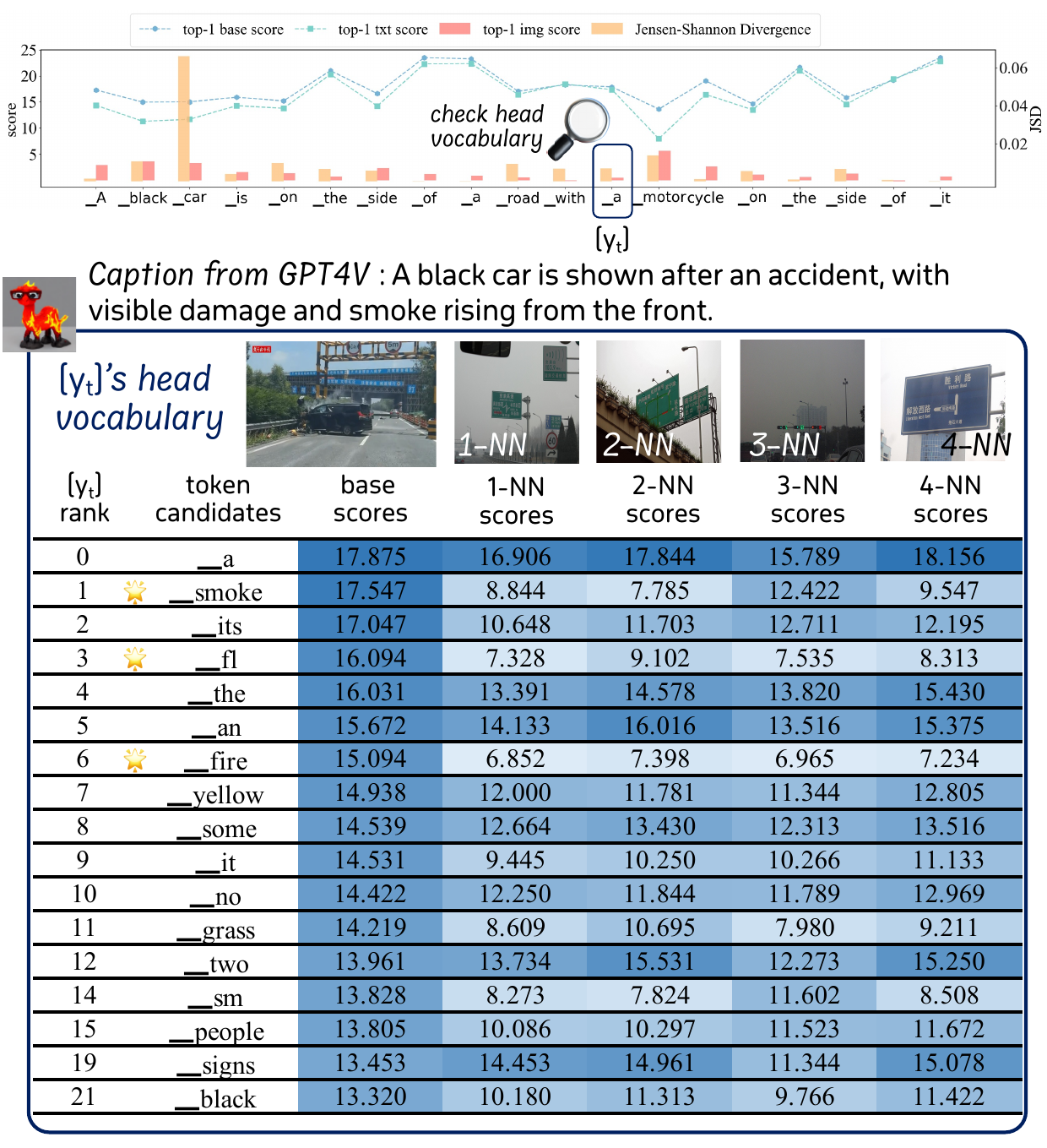}
  \caption{
    More qualitative evidence to support our hypothesis.
    This caption is predicted by LLaVA1.5-7B.
    The test image is crawled from the internet,
    and the references are from the COCO Caption dataset.
    The star emojis indicate the accurate candidates.
  }
  \label{fig:more_evidence_web02}
\end{figure}

\begin{figure}[tb]
  \centering
  \includegraphics[height=16.8cm]{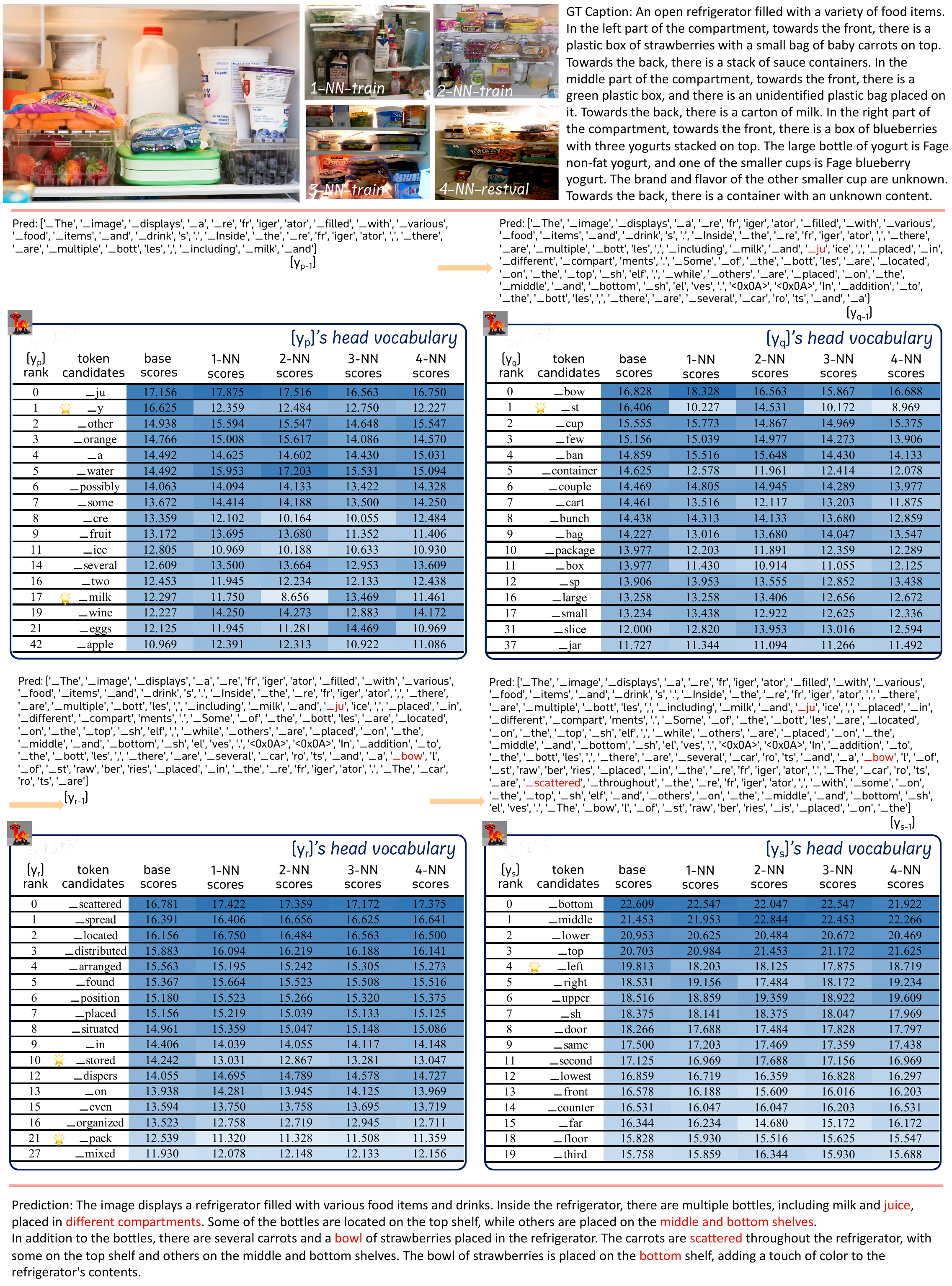}
  \caption{
    More qualitative evidence to support our hypothesis.
    This sample is from the LLaVA Bench in the wild~\cite{liu2024visual} benchmark.
    Hallucinatory content is marked red.
  }
  \label{fig:more_evidence_lbench}
\end{figure}

We provide more examples in Figures~\cref{fig:more_evidence_whoops01} to~\cref{fig:more_evidence_lbench},
with test images from the Whoops
benchmark,
LLaVA Bench in the wild,
and crawled from the internet.
These examples further validates our three main claims:

\begin{itemize}
    \item \textbf{MLLMs are not completely ignorant of the accurate visual cues amidst visual hallucination.}
    In~\cref{fig:more_evidence_whoops01} to~\cref{fig:more_evidence_lbench},
    we notice that
    at least one accurate candidate consistently ranks among the top-3 most probable candidates in $\mathcal V_{head}^{m}$,
    with comparable confidence score to the top-ranked candidates,
    \eg, $\_hot$ gets 15.406 scores and ranks second (-0.032 compared to top one) in~\cref{fig:more_evidence_whoops01},
    $\_rub$ scores 19.203 and ranks second (-0.375 compared to top one) in~\cref{fig:more_evidence_whoops02},
    $\_fuel$ scores 13.695 and ranks second (-0.304 compared to top one) in~\cref{fig:more_evidence_web01},
    $\_smoke$ scores 17.547 and ranks second (-0.328 compared to top one) in~\cref{fig:more_evidence_web02},
    $\_y$ (first token for yogurt) and $\_st$ (first token for strawberry) 
    score 16.625 (-0.531 compared to top one) and 16.406 (-0.422 compared to top one), respectively, in~\cref{fig:more_evidence_lbench}.
    In other words,
    the hallucinatory candidates marginally exceed the faithful ones
    and lead to visual hallucinations in the responses.
    This observation validates the feasibility of
    mitigating visual hallucinations
    by moderately adjusting the predicted confidence score distribution.

    \item \textbf{Similar Images can Induce Analogous Visual Hallucination.}
    In~\cref{fig:more_evidence_whoops01} to~\cref{fig:more_evidence_lbench},
    we observe that in the same context,
    analogous visual hallucination can occur among
    images with similar semantic and appearance characteristics.
    Specifically, hallucinatory candidates may get one or more high 
    \textit{kNN score}
    even though they can \textit{\textbf{not}} be grounded in the corresponding \textit{kNN} image.
    For instance,
    in~\cref{fig:more_evidence_whoops01},
    candidate $\_c$ and $\_l$ (first token for lollipop) have more than 14 confidence scores
    in the third and fourth visual references that contain only children and toothbrushes.
    In~\cref{fig:more_evidence_whoops02},
    candidate $\_du$ scores 20.063 and 19.838 
    in the first and third references, respectively, which contain swans only.
    In~\cref{fig:more_evidence_web01},
    candidate $\_fire$ scores more than 10 in images without fire-related content,
    and $\_person$ scores 12.984 and 12.180 in images without people,
    which is very close to the predicted scores when the input images do contain people (13.344 and 13.188).
    In~\cref{fig:more_evidence_web02},
    the ambiguous candidate $\_two$ scores from 12.273 to 15.531
    across images crowded with multiple objects.
    In~\cref{fig:more_evidence_lbench},
    candidates $\_ju$ (first token for juice),
    $\_orange$,
    and $\_bow$
    have high confidence scores across all reference images,
    yet some of the references do not contain such visual content.
    Note that we do not deny that
    candidates with high \textit{kNN scores} across all references
    can be the correct content that the references have in common,
    we pinpoint our observation that they 
    are more likely to
    be hallucinations
    compared to candidates with significant score changes.

    \item \textbf{Visual References Help Discern Accurate Content.}
    Compared to the
    candidates with subtle confidence score changes,
    we claim that
    candidates with more significant score variation
    are \textit{\textbf{more likely}} to be the accurate ones.
    For instance,
    in~\cref{fig:more_evidence_whoops01},
    the accurate candidate $\_hot$'s score changed from 15.406 to 11.428 in average,
    where as the hallucinatory candidate $\_c$'s score changed from 15.438 to 14.820.
    in~\cref{fig:more_evidence_whoops02},
    the accurate candidate $\_rub$'s score changed from 19.203 to 8.857 in average,
    which is much more drastic than that of other hallucinatory 
    (\eg, $\_du$ from 19.578 to 17.609) 
    or ambiguous ($\_baby$ from 18.984 to 15.943) 
    candidates.
    In~\cref{fig:more_evidence_web01},~\cref{fig:more_evidence_web02},
    and~\ref{fig:more_evidence_lbench},
    we also observe that the \textit{kNN scores} of all accurate candidates
    (marked with a star emoji)
    are significantly \textit{\textbf{lower}} than that of other erroneous candidates
    with close ranks in the vocabulary,
    \eg, $\_fuel$, $\_gas$, and $\_tank$ in~\cref{fig:more_evidence_web01},
    $\_smoke$, $\_fl$ (first token for flame), and $\_fire$ in~\cref{fig:more_evidence_web02},
    as well as $\_y$, $\_st$, and $\_left$ in~\cref{fig:more_evidence_lbench}.
    This phenomenon could be leveraged to discern the accurate content from hallucinations.
    In practice, we contrast the scores predicted by the test images
    to other scores predicted by the retrieved similar images,
    thereby promoting the candidates that are more likely to be the faithful ones.
    Note that we do \textbf{\textit{not}} pre-establish or assume certain candidates
    to be the correct ones,
    Instead, we involve all top-50 ranked candidates in $\mathcal V_{head}^{m}$
    for score comparison,
    facilitating MLLMs to discern \textbf{\textit{multiple potentially correct}} candidates
    by comparing similar but non-identical images.
    
\end{itemize}

\end{document}